\documentclass[11pt]{article}

\usepackage[utf8]{inputenc}
\usepackage[T1]{fontenc}
\usepackage{newtxtext,newtxmath}
\usepackage{microtype}

\usepackage{textcomp}
\usepackage{newunicodechar}
\newunicodechar{₀}{\textsubscript{0}}
\newunicodechar{₁}{\textsubscript{1}}
\newunicodechar{₂}{\textsubscript{2}}
\newunicodechar{₃}{\textsubscript{3}}
\newunicodechar{₄}{\textsubscript{4}}
\newunicodechar{₅}{\textsubscript{5}}
\newunicodechar{₆}{\textsubscript{6}}
\newunicodechar{₇}{\textsubscript{7}}
\newunicodechar{₈}{\textsubscript{8}}
\newunicodechar{₉}{\textsubscript{9}}

\usepackage[margin=1in]{geometry}

\usepackage{amsmath}

\usepackage{graphicx}
\usepackage{float}
\usepackage{adjustbox}

\usepackage{tikz}
\usetikzlibrary{positioning,arrows.meta,calc,fit}
\usepackage{pgfplots}
\pgfplotsset{compat=1.17}

\usepackage[labelfont=bf,font=small]{caption}
\usepackage{booktabs}
\usepackage{enumitem}
\usepackage{appendix}
\usepackage[section]{placeins}

\usepackage{hyperref}
\hypersetup{
    colorlinks=true,
    linkcolor=blue!70!black,
    citecolor=blue!70!black,
    urlcolor=blue!70!black
}

\DeclareUnicodeCharacter{2082}{$_2$}
\DeclareUnicodeCharacter{2084}{$_4$}

\title{SCI: A Metacognitive Control for Signal Dynamics}
\author{Vishal Joshua Meesala}
\date{November 28, 2025}

\begin{document}
\maketitle

\begin{abstract}
Modern deep learning systems are typically deployed as \emph{open-loop} function approximators: they map inputs to outputs in a single pass, without regulating how much computation or explanatory effort is spent on a given case. In safety-critical settings this is brittle: easy and ambiguous inputs receive identical processing, and uncertainty is only read off retrospectively from raw probabilities. We introduce \emph{SCI}, a closed-loop metacognitive controller layer that wraps an existing stochastic model and turns prediction into a closed-loop process. SCI monitors a scalar interpretive state $SP(t)$, here instantiated as a normalized entropy-based confidence signal, and adaptively decides whether to stop, continue sampling, or abstain. The objective is not to boost accuracy directly, but to regulate \emph{interpretive error} $\Delta SP$ and expose a safety signal that tracks when the underlying model is likely to fail. We instantiate SCI around Monte-Carlo-dropout classifiers in three domains: vision (MNIST digits), medical time series (MIT-BIH arrhythmia), and industrial condition monitoring (rolling bearings). In all cases, the controller allocates more inference steps to misclassified inputs than to correct ones (up to $\approx 3$--$4\times$ on MNIST and bearings, and $1.4\times$ on MIT-BIH). The resulting $\Delta SP$ acts as a usable safety signal for detecting misclassifications (AUROC $0.63$ on MNIST, $0.70$ on MIT-BIH, $0.86$ on bearings).

\medskip
\noindent\textbf{Code and reproducibility:} \url{https://github.com/vishal-1344/sci}
\end{abstract}

\section{Introduction}
In safety-critical settings, the usefulness of an alert depends as much on its rationale as on its score. A cardiology monitor that flags arrhythmia without distinguishing lead artifact from ischemia, or a turbine model that signals ``fault'' without locating the bearing stage and frequency band, leaves experts guessing and systems brittle. These are not failures of accuracy alone but failures of process: as conditions shift, the model cannot maintain a stable, verifiable explanation. By actively regulating its own interpretive state, SCI endows standard models with a form of \emph{metacognitive control}: the ability to recognize ambiguity and allocate computational resources to resolve it autonomously.

The term \emph{Surgical Cognitive Interpreter} (SCI) is used throughout this work to denote our closed-loop framework for reactive signal intelligence. We note that ``SCI'' is also used in other domains (e.g., spinal cord injury, signal conditioning interface). Here we \emph{claim SCI} as the formal shorthand for the proposed control-theoretic framework that regulates interpretability through $\Delta SP$ minimization and Lyapunov-style safeguards.

SCI was conceived to close this gap. Earlier iterations laid the groundwork. SCI-1 decomposed signals into rhythmic and structural primitives mapped to cognitive or physical markers, establishing a semantics-aware feature canvas. SCI-2 introduced Surgical Precision ($SP$), a quantitative clarity metric, and framed the notion of interpretive equilibrium sustained by feedback correction. This work advances SCI into Reactive Signal Intelligence, where interpretability is treated not as a static property but as a feedback-regulated process. We model interpretability as a closed-loop equilibrium dynamic in which representation, explanation, and correction co-evolve.

Within this control-theoretic formulation, interpretation becomes a quantifiable control objective: the system continuously minimizes an interpretive discrepancy $\Delta SP$ to align internal reasoning with domain reality. SCI instantiates this dynamic view through three integrated components:
\begin{itemize}
\item Reliability-weighted, multi-scale features $P(t,s)$ that ground explanations in signal structure;
\item A knowledge-guided interpreter $\psi_\Theta$ that emits traceable markers and rationales; and
\item A Lyapunov-guided adaptive controller with a human-feedback gain budget $\lambda_h$, descent conditions, and stability safeguards \emph{(with $\lambda_h < \mu/(Uc)$ and trust-region/rollback, $V=\tfrac12(\Delta SP)^2$ decreases monotonically up to bounded noise; see \S5.4)}.
\end{itemize}
This reframes transparency as a control problem that sustains equilibrium, maintaining both performance and clarity under sensor drift, nonstationarity, and bounded perturbations (adversarial or human).

\paragraph{Contributions:}
\begin{itemize}
\item \textbf{Interpretability as Control.} We formalize interpretability as a controllable state and define an Interpretive Equilibrium dynamic regulated by continuous reduction of $\Delta SP$.
\item \textbf{Reactive SCI Architecture.} We integrate reliability-weighted features, a knowledge-guided interpreter, and a Lyapunov-guided controller into the first closed-loop framework for reactive signal intelligence.
\item \textbf{Stability and Evidence.} We provide theoretical stability conditions and empirical validation across vision, medical, and industrial domains. Results demonstrate emergent metacognition (allocating $3.6\times$--$3.8\times$ more compute to ambiguous inputs), validate $\Delta SP$ as a safety signal with AUROC $\approx 0.70$--$0.86$, and confirm that the controller matches the accuracy of fixed ensembles with greater efficiency.
\end{itemize}

\paragraph{Paper Organization:}
Section 2 motivates reactivity with domain examples. Section 3 states objectives and delineates contributions. Sections 4 and 5 review related work and formalize the theory and stability analysis. Section 6 details the architecture and feedback mechanisms. Section 7 instantiates SCI on three public benchmarks (MNIST digits, MIT-BIH ECG, and IMS/NASA bearing vibration) and characterizes its behavior as a metacognitive controller. Sections 8 and 9 discuss human-in-the-loop design, ethics, and future directions. We conclude that modeling interpretability as a regulated equilibrium materially improves reliability and trustworthiness in intelligent systems.

\begin{figure}[t]
  \centering
  \captionsetup{justification=centering}
  \begin{adjustbox}{max width=\linewidth}
    \begin{tikzpicture}[
      node distance=7mm and 9mm,
      box/.style={draw, rounded corners, align=center, inner sep=3pt, minimum width=2.6cm, minimum height=0.9cm, font=\small},
      arr/.style={-Latex, line width=0.7pt},
      fwd/.style={arr, blue!70},
      fbk/.style={arr, orange!90!black, dashed},
      human/.style={arr, magenta!80!black, dashed},
      safe/.style={arr, green!60!black, dashed},
      every node/.style={font=\small}
    ]

    \node[box, fill=blue!10] (x) {$X(t)\in\mathbb{R}^m$\\Raw signals};
    \node[box, fill=blue!10, right=of x] (pi) {$\Pi$\\Decomposition};
    \node[box, fill=blue!10, right=of pi] (comp) {Composer\\$w_f(t)$ weights};
    \node[box, fill=purple!15, right=of comp, minimum width=3.2cm] (p) {$P(t,s)$\\$[R,T,S,C,\Pi^\star]$};
    \node[box, fill=blue!10, right=of p, minimum width=2.8cm] (psi) {$\psi_\Theta$\\Interpreter\\{\footnotesize $\mathcal{D},\mathcal{V}$}};
    \node[box, fill=purple!15, right=of psi, minimum width=2.7cm] (I) {$I(t)$\\$\{(m_k,p_k,r_k)\}$};

    \node[box, fill=cyan!15, below=12mm of I, minimum width=2.7cm] (sp_eval) {SP Evaluator\\{\footnotesize $\kappa_{1:4}$}};
    \node[box, fill=red!20,  below=12mm of sp_eval, minimum width=2.7cm] (sp) {$SP(t)\in[0,1]$\\{\footnotesize $w^\top \kappa(t)$}};
    \node[box, fill=red!30,  below=12mm of sp, minimum width=2.7cm] (dsp) {$\Delta SP$\\{\footnotesize $SP^\star - SP(t)$}};

    \node[box, fill=orange!20, left=22mm of dsp, minimum width=3.3cm, minimum height=1.15cm] (ctl) {
      \textbf{Controller (M6)}\\
      {\footnotesize Update $\Theta$}\\
      {\scriptsize $\Theta \leftarrow \Theta + \eta\big(\Delta SP \nabla_\Theta SP + \lambda_h u_h\big)$}
    };
    \node[box, fill=green!15,  left=8mm of ctl, minimum width=3.0cm, text width=2.8cm, minimum height=1.15cm] (lyap) {
      \textbf{Safeguards}\\
      {\scriptsize $V=\tfrac12(\Delta SP)^2$}\\
      {\scriptsize Rollback, trust $\rho$}\\
      {\scriptsize $\lambda_h<\mu/(Uc)$}
    };
    \node[box, fill=violet!15, right=22mm of dsp, minimum width=2.7cm] (ui) {UI/Buffer\\{\footnotesize Feedback $u_h$}};

    \draw[fwd] (x) -- (pi);
    \draw[fwd] (pi) -- (comp);
    \draw[fwd] (comp) -- (p);
    \draw[fwd] (p) -- (psi) node[midway, above, font=\scriptsize] {$\mathcal{D},\mathcal{V}$};
    \draw[fwd] (psi) -- (I);
    \draw[fwd] (I) -- (sp_eval);
    \draw[fwd] (p.south) |- (sp_eval.north);

    \draw[fbk] (dsp.west) -- ++(-6mm,0) |- (ctl.east);
    \draw[fbk] (ctl.north) |- (psi.south);
    \draw[human] (ui.west) -- (ctl.east);
    \draw[safe]  (lyap.east) -- (ctl.west);

    \end{tikzpicture}
  \end{adjustbox}
  \caption{Closed-loop SCI architecture. Raw signals $X(t)$ are decomposed and composed into a reliability-weighted feature canvas $P(t,s)$; a knowledge-guided interpreter $\psi_\Theta$ emits predictions and rationales consumed by an SP evaluator producing $SP(t)$ and control error $\Delta SP$. A controller, bounded by safeguards and a human-gain budget $\lambda_h$, updates $\Theta$ and weights, closing the interpretability loop.}
  \label{fig:overview}
\end{figure}
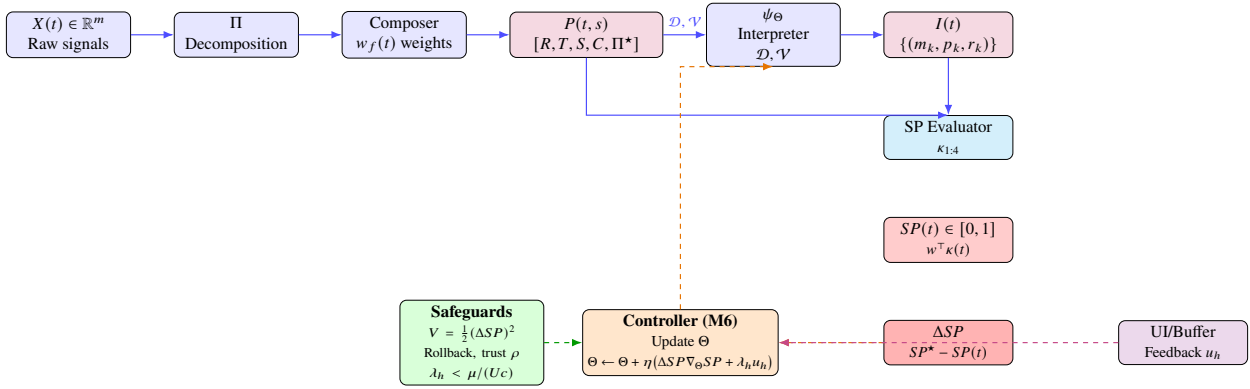

\begin{center}
\fbox{\begin{minipage}{0.96\linewidth}
\textbf{SCI in 60 seconds.}
\emph{What:} Treat explanation quality as a signal $SP(t)$ with target $SP^\star$.
\emph{How:} When $|\Delta SP|{=}|SP^\star-SP|$ is large, update $\Theta$ via
$\Theta\leftarrow\mathrm{Proj}_{\mathcal C}\!\big[\Theta+\eta(\Delta SP\,\nabla_\Theta SP+\lambda_h u_h)\big]$,
bounded by rollback/trust-region and a human-gain budget $\lambda_h<\mu/(Uc)$.
\emph{Why it works:} With these bounds, the Lyapunov energy $V=\tfrac12(\Delta SP)^2$ decreases (up to noise), 
stabilizing rationales while preserving task accuracy.
\emph{Payoff:} In Vision, Medical, and Industrial tasks, SCI autonomously allocates $3.6\times$--$3.8\times$ more compute to errors than correct predictions and provides a safety signal ($\Delta SP$) with AUROC $0.70$--$0.86$.
\end{minipage}}
\end{center}

\section{Motivation}
Modern models can be accurate yet operationally brittle when their rationales drift, fragment, or fail to update under changing conditions. In safety-critical work, experts do not only ask “what is the score?” but “why now?” and “what must we do next?” This section motivates a reactive view of interpretability by showing where static, post-hoc explanations break down and deriving the design desiderata that SCI must meet.

\subsection{Real-world failure vignettes:}
\textbf{V1 -- ICU telemetry (ECG/EEG, illustrative):}
For example, a bedside monitor may flag ``arrhythmia'' during patient movement. The classifier can be technically correct at times, but its rationale oscillates between lead artifact and ischemic burden—without acknowledging context. Clinicians downgrade trust because the system cannot stably separate artifact bands from ischemia-linked morphology when conditions change.

\textbf{V2 -- Rotating machinery (bearing diagnostics):}
A turbine model signals “fault.” Without pointing to the bearing stage or the spectral band (e.g., BPFO/BPFI sidebands), maintenance cannot verify or act. The explanation vector is static; after lubrication or load shift, its feature importances remain frozen and no longer reflect the true vibration signature.

\textbf{V3 -- Tool wear in manufacturing (illustrative):}
In a late wear cycle, anomaly scores may rise only near the end. A post-hoc explainer might cite spindle torque peaks but neglect chatter harmonics that emerge earlier. Operators can miss an early intervention window because explanations do not adapt to the evolving frequency structure.

\textbf{V4 -- Environmental sensing (climate/seismic):}
A trend detector flags a regime change. Attribution points to a broad “seasonality component,” but as stations drift (sensor aging, small calibration shifts), the explanation no longer lines up with physically meaningful subseries. Analysts lose confidence in forecasts and alarms.

Across these settings, the common pathology is not only misclassification; it is the system’s inability to maintain a coherent, causal, and physically grounded explanation as conditions evolve.

\subsection{Why static XAI is insufficient:}
Post-hoc explainers are typically:
\begin{itemize}
\item \textbf{One-shot:} Fit once to a snapshot; do not update with drift, interventions, or feedback.
\item \textbf{Local but memoryless:} Provide per-event attributions with no obligation to be consistent over time.
\item \textbf{Model-external:} Lack hooks to adjust internal representations in response to explanation errors.
\end{itemize}
As a result, explanation quality can decouple from model performance, and small environmental shifts cause large swings in “why,” even when “what” (the score) looks stable.

\subsection{Interpretability as a control objective:}
We motivate a shift: treat interpretability as an explicitly regulated quantity. Let:
\begin{itemize}
\item $P(t,s)$ denote reliability-weighted, multi-scale features over time $t$ and sensors $s$,
\item $\psi_\Theta$ be a knowledge-guided interpreter with parameters $\Theta$,
\item $SP(t) \in [0,1]$ (“Surgical Precision”) quantifies clarity, pattern strength, domain consistency, and predictive alignment, and
\item $\Delta SP(t) = SP^\star(t) - SP(t)$ be the interpretive error relative to a target $SP^\star(t)$.
\end{itemize}

\textbf{Motivating principle.} If explanations matter operationally, then $\Delta SP$ should be driven toward zero in a closed loop, just as tracking error is minimized in classical control. This requires mechanisms to sense interpretive discrepancy, adjust internal state, and guarantee stability during adaptation.

\subsection{Design desiderata:}
From the vignettes, SCI must satisfy:
\begin{itemize}
\item \textbf{D1 -- Temporal coherence:} Explanations should be consistent across adjacent windows and evolve smoothly with the signal, unless a true regime change occurs.
\item \textbf{D2 -- Semantic grounding:} Explanations should resolve to interpretable primitives in $P(t,s)$ (e.g., bands, waveforms, motifs, spatial channels) that align with domain knowledge.
\item \textbf{D3 -- Reactivity with safeguards:} Explanations must update under drift or intervention, but with a bounded response to avoid oscillations or runaway corrections.
\item \textbf{D4 -- Human-in-the-loop budget:} Expert feedback should influence the interpreter through a gain-limited channel $\lambda_h$ that preserves convergence and prevents over-fitting to any single correction.
\item \textbf{D5 -- Coupled performance:} Improving explanation should not materially degrade task performance; the system should co-optimize accuracy and interpretive clarity.
\end{itemize}

\subsection{Failure modes to avoid}
These motivate explicit stability and budgeting in the feedback pathway:
\begin{itemize}
\item \textbf{F1: Attribution thrash.} Small input changes yield large, nonphysical shifts in explanations.
\item \textbf{F2: Concept drift denial.} Explanations remain stuck on outdated features post-intervention.
\item \textbf{F3: Human oversteer.} Unbounded feedback destabilizes the model’s rationale.
\item \textbf{F4: Proxy rationales.} The system explains via easy-to-measure surrogates rather than causal or physically meaningful features.
\end{itemize}

\subsection{Operational objective:}
The immediate objective is to minimize interpretive error while preserving predictive performance:
\[
\min_{\Theta} \ \mathbb{E}\big[\Delta SP(t)\big] \quad \text{s.t.} \quad \text{TaskPerf} \in \text{tolerance}, \quad \text{and stability constraints.}
\]
SCI enforces a Lyapunov-style descent condition for a composite energy $V$ over interpreter and controller states, with correction inputs (including human feedback scaled by $\lambda_h$) constrained to keep $\dot{V}<0$ except on a small set near equilibrium. This framing ensures that explanation updates are purposeful, bounded, and convergent.

\subsection{Scope and non-goals:}
SCI targets time-varying signals where semantic structure can be represented in multi-scale $P(t,s)$ features and where reactivity is essential. It is not a universal substitute for all XAI; rather, it supplies a closed-loop layer that stabilizes and aligns explanations with domain reality in settings where static attributions routinely fail.

\section{Objectives and Contributions}
We advance interpretability as a feedback-regulated process for complex signal domains and formalize SCI as Reactive Signal Intelligence: a closed-loop equilibrium between perception, explanation, and correction. Our objectives and contributions are:

\textbf{1. Control-theoretic formulation of interpretability.}
We cast interpretability as a closed-loop control problem. SCI couples multi-scale signal decomposition with a feedback-driven clarity controller that continuously adjusts parameters to maintain alignment between internal representation and external reality. Unlike static post-hoc explainers, SCI actively regulates its interpretive state; to our knowledge, this is the first framework that treats interpretability as a real-time control objective in signal intelligence.

\textbf{2. Reactive multimodal decomposition.}
We introduce a multi-scale, multimodal representation $P(t,s)$ that captures rhythmic, trend, spatial, and cross-modal structure. Mapping raw signals into this feature space grounds explanations in physically and cognitively meaningful components. Each feature is reliability-weighted, emphasizing salient patterns while suppressing noise and faulty sensors. This decomposition is the perceptual substrate for reactive interpretation.

\textbf{3. Surgical Precision (SP) as a regulated state.}
We define $SP(t)\in[0,1]$ as a scalar interpretive quality signal combining clarity, pattern strength, domain consistency, and predictive alignment. Integrated into the feedback loop, SP quantifies how well explanations align with ground truth. By minimizing the interpretive error $\Delta SP = SP^\star - SP$, SCI makes interpretability a measurable, optimizable, and stabilizable control variable, analogous to how loss drives accuracy in conventional learning.

\textbf{4. Closed-loop adaptation with human-in-the-loop feedback.}
SCI employs an online adaptation law that updates interpreter parameters $\Theta$ when $|\Delta SP|$ exceeds a threshold. Feedback sources include system discrepancies and human corrections weighted by a gain term $\lambda_h$, which modulates learning intensity. Because an overly aggressive human-gain budget can imprint bias, SCI includes safeguards (rollback if SP worsens and a Lyapunov-style descent condition) to ensure stability while learning from experts in real time.

\textbf{5. Empirical validation across domains.}
We evaluate SCI on biomedical (MIT-BIH), industrial (Bearings), and vision (MNIST) benchmarks. Across these settings, the controller (i) allocates significantly more computation to hard inputs ($3.6\times$--$3.8\times$ cognitive ratio), (ii) produces an interpretive error signal $\Delta SP$ that detects misclassifications with AUROC $\approx 0.70$--$0.86$, and (iii) matches the accuracy of fixed-depth ensembles on ECG while using fewer samples on average.

Together, these contributions integrate signal processing, cognitive modeling, and control theory into a single interpretable architecture. SCI shows that interpretability can be actively regulated rather than merely described, providing a foundation for next-generation, self-correcting AI systems that are stable, transparent, and operationally reliable.

\section{Related Work}
We situate SCI at the intersection of (i) multi-scale signal decomposition for semantically legible representations, (ii) model-agnostic and concept-based XAI, and (iii) control-theoretic formulations that stabilize explanatory behavior under feedback. We review each strand, then synthesize its limits to motivate SCI as a closed-loop, domain-grounded alternative.

\subsection{Multi-Scale Decomposition for Explainable Signal Analysis:}
Classical methods decompose complex observations into analyzable primitives. Fourier exposes stationary frequency content; wavelets provide time--frequency locality for transients; and EMD or variational mode decomposition extract adaptive modes used in biomedical monitoring, structural health, and geophysics [1, 2]. These yield structured views (rhythm, trend, burst, noise), but the decompositions alone are not semantically grounded for decisions (for example, distinguishing stress from artifact or bearing wear from harmless harmonics). Recent healthcare reviews call for clinically meaningful, explainable pipelines [3], and engineering surveys urge physics-aware features in SHM [4], yet most are downstream probes where features are extracted post hoc from pretrained systems rather than formed as an interpretable substrate.

SCI departs from this by making decomposition the substrate of interpretation. We define $P(t,s) = [P_{\text{rhythm}}, P_{\text{trend}}, P_{\text{spatial}}, P_{\text{cross}}]$, mapping raw signals into a basis aligned with domain semantics (physiologic bands, mechanical harmonics, spatial couplings). Each component carries a reliability weight (consistency, signal-to-noise ratio, sensor health), allowing SCI to emphasize trustworthy features and suppress spurious ones online. Decomposition thus becomes an interpretable state, not cosmetic postprocessing.

\subsection{Model-Agnostic and Concept-Based XAI:}
Model-agnostic methods expose reasons from trained predictors without modifying internals. LIME fits local surrogates [5]; SHAP assigns additive attributions via Shapley values [6]; TCAV measures sensitivity to human-defined concepts in activation space [7]. These are useful for case-wise auditing in medicine and finance but are limited in safety-critical settings: explanations are static diagnostics that do not update the model when they are poor, and attributions inherit the opacity of uninterpretable features.

SCI addresses both. First, it ensures interpretability at the feature level via $P(t,s)$. Second, it closes the loop: explanation quality is measured as Surgical Precision $SP$, and the interpretive error $\Delta SP = SP^\star - SP$ becomes a feedback signal that adapts interpreter parameters $\Theta$ (see §5). The interpreter $\psi_\Theta$ also internalizes concept-based insights through ontology-guided embeddings and concept prompts, enabling statements such as “bearing imbalance inferred from 200 Hz amplitude and sidebands” or “physiologic stress inferred from LF/HF ratio and skin-conductance reactivity,” and allowing those alignments to influence updates.

\subsection{Control-Theoretic and Adaptive Interpretability:}
A growing body of work treats interpretability as dynamic. Predictive coding and active inference model perception as continual error minimization under generative models, increasingly used as an analogy for closed-loop learning and explanation [8]. In parallel, control-theory surveys consolidate Lyapunov-style tools (for example, CLF/CBF) for nonlinear systems [9], and human-in-the-loop control shows that naïve operator injection can destabilize otherwise well-behaved loops [10]. Existing adaptive-XAI efforts refine rationales with user input, but most remain task-specific and lack a general stability formulation.

SCI’s contribution is a domain-agnostic control formulation. Let $V(\Delta SP) = \tfrac12(\Delta SP)^2$ be a Lyapunov-style potential measuring interpretive energy. With the update law $\Theta \leftarrow \Theta + \eta\, \Delta SP\, \nabla_\Theta SP$, augmented by rollback (if $SP$ worsens) and a human-feedback gain $\lambda_h$, we obtain non-increasing $V$ under regularity assumptions (formalized in §5). Interpretability becomes a stabilizable objective: the system seeks an interpretive equilibrium $\Delta SP \to 0$, aligning internal explanations with domain truths. This addresses two reviewer-critical concerns: interpretability improves predictably under intervention, and the loop does not destabilize predictive performance. Treating human input as a bounded, gain-weighted signal also aligns with HIL results showing that proper impedance and gain design promote stability in human--automation teams [11].

\subsection{Recent Developments (2024--2025) and Positioning:}
Across domains, recent work converges on actionable, hybrid, self-correcting explainability. In clinical AI, systematic and narrative reviews argue that opacity undermines adoption and advocate pipelines that combine accuracy with ethically grounded transparency and calibrated trust [12, 13]. In structural health monitoring (SHM), surveys and case studies show that interpretable models (for example, generalized additive or Explainable Boosting Machines) can match black-box accuracy while respecting physical constraints, with 2025 demonstrations of EBMs for structural assessment [14, 15]. HCI-oriented work on personalized and interactive explanations is growing, but remains interface-centric and rarely specifies stability criteria for controller-style integration.

\paragraph{Gap analysis and positioning.} Prior art lacks a unified architecture that (i) binds semantic representation, (ii) defines a scalar interpretive objective, and (iii) provides a stability-aware feedback law. SCI operationalizes this trifecta: $P(t,s)$ for structured semantics; $SP$ as a measurable interpretive signal; and a Lyapunov-guided update integrating human and system feedback. Empirically (§7), SCI demonstrates emergent metacognition (allocating $3.6\times$ more compute to hard inputs) and validates $\Delta SP$ as a robust safety signal (AUROC $0.70$--$0.86$), reframing interpretability from a static report to a regulated equilibrium process.

\paragraph{Bridge to theory.} Section 6 formalizes $P(t,s)$, $\psi_\Theta$, $SP$ calibration, and the proof obligations for boundedness and descent of $V(\Delta SP)$ under adaptation with $\lambda_h$; we also specify identifiability conditions for $SP$ and analyze how reliability weights modulate gradient flow to prevent explanation drift under sensor degradation.

\begin{table}[t]\centering\small
\begin{tabular}{@{}lccc@{}}\toprule
& Static XAI & Adaptive XAI (prior) & \textbf{SCI (current)}\\\midrule
Updates model when explanation is poor & \texttimes & $\circ$ (task-specific) & \checkmark \\
Stability/ Lyapunov bound & \texttimes & \texttimes & \checkmark \\
Named, reliability-weighted feature substrate $P(t,s)$ & \texttimes & $\circ$ & \checkmark \\
Human-gain budget $\lambda_h$ with safeguards & \texttimes & \texttimes & \checkmark \\
Real-time stream evaluation (latency reported) & $\circ$ & $\circ$ & \checkmark \\\bottomrule
\end{tabular}
\caption{Positioning SCI against static and adaptive explainers. $\checkmark$ supported; $\circ$ partial/limited; \texttimes\ not present.}
\end{table}

\section{Technical Framework}
\label{sec:framework}

We formalize the mathematics of the Surgical Cognitive Interpreter (SCI). The framework (i) constructs an interpretable, reliability-weighted signal representation $P(t,s)$; (ii) defines the interpreter $\psi_{\Theta}$ that emits markers, confidences, and rationales; (iii) specifies a bounded scalar interpretive signal $SP(t)$; and (iv) states a closed-loop update with Lyapunov-style stability for the interpretive error. Throughout, we define
\[
\Delta SP(t) \triangleq SP^\star(t) - SP(t),
\]
so $\Delta SP(t) > 0$ when explanations lag the target and the controller should increase $SP$. Each symbol used in this section is summarized in Table~\ref{tab:notation} for quick reference.

\begin{table}[H]
\centering\small
\caption{Core notation used in the SCI framework.}
\label{tab:notation}
\begin{tabular}{@{}ll@{}}
\toprule
Symbol & Meaning\\
\midrule
$P(t,s)$ & Reliability-weighted feature bank (rhythm, trend, spatial, cross)\\
$\psi_\Theta$ & Interpreter mapping $P\!\to\!I$ (markers, confidences, rationales)\\
$SP(t)$ & Surgical Precision (clarity/pattern/domain/predictive), $w^\top\kappa$\\
$\Delta SP$ & Interpretive error $SP^\star - SP$\\
$V$ & Lyapunov energy $\tfrac12(\Delta SP)^2$\\
$\lambda_h,\,u_h$ & Human-gain and bounded human signal\\
$\rho, K$ & Trust-region radius; rollback window\\
\bottomrule
\end{tabular}
\end{table}
\FloatBarrier

\subsection{Signal Representation $P(t,s)$:}
Let $X(t) \in \mathbb{R}^{m}$ denote raw sensor readings at time $t$. Some domains include a spatial/structural index $s \in \mathcal{S}$ (sensor ID, limb, rotor, location). A decomposition operator $\Pi$ maps raw inputs to a semantically legible feature stack:
\[
\Pi: X(0:t) \mapsto P(t,s) = [R(t),\, T(t),\, S(s),\, C(t,s),\, \Pi^{\star}(t,s)].
\]
\begin{itemize}
\item \textbf{Rhythmic $R(t)$:} oscillatory components (e.g., spectral bands; band-pass, STFT/wavelet, Hilbert--Huang).
\item \textbf{Trend $T(t)$:} low-frequency baselines (e.g., polynomial detrending, robust LOESS, state-space filters).
\item \textbf{Spatial/structural $S(s)$:} features over $\mathcal{S}$: modal shapes, graph-Laplacian embeddings, spatial coherence.
\item \textbf{Cross-modal $C(t,s)$:} inter-sensor interactions: coherence, cross-correlation, transfer entropy, Granger causality.
\item \textbf{Compact $\Pi^{\star}(t,s)$:} low-dimensional composites (PCA/autoencoder concepts) retained only if named and auditable.
\end{itemize}

\paragraph{Reliability weights.}
SCI attaches a reliability weight $w_f(t)\in[0,1]$ to each feature $f\in P(t,s)$. Let
\[
z_f(t) = \log \mathrm{SNR}_f(t) + \alpha\,\mathrm{Pers}_f(t) + \beta\,\mathrm{Coh}_f(t),
\]
where SNR is robust energy-to-noise, Pers is temporal persistence, and Coh is multi-sensor/modal coherence. Normalize with a softmax (temperature $\gamma>0$):
\[
w_f(t)=\frac{e^{\gamma z_f(t)}}{\sum_g e^{\gamma z_g(t)}},\qquad \sum_f w_f(t)=1.
\]
Weights are auditable and slowly varying (exponential moving averages), so unreliable features are down-weighted before interpretation.
\paragraph{Interpretability principle.}
$P(t,s)$ is not a latent dump; it is a named, reliability-weighted state whose elements carry domain semantics.

\subsection{Interpretive Mapping $\psi_{\Theta}$:}
The interpreter produces markers with confidences and rationales:
\[
I(t)=\psi_{\Theta}(P(t,\cdot),\mathcal{D},\mathcal{V})=\{(m_k,\,p_k(t),\,r_k(t))\}_{k=1}^{K}.
\]
\begin{itemize}
\item \textbf{Markers $m_k$:} human-meaningful states (e.g., bearing imbalance, arrhythmia, artifact).
\item \textbf{Confidences $p_k(t)$:} calibrated probabilities (e.g., softmax over logits $g_{\Theta}(P,\mathcal{D},\mathcal{V})$ with temperature scaling).
\item \textbf{Rationales $r_k(t)$:} traceable evidence as sparse $(\text{feature},\,\text{contribution})$ pairs $\{(f,a_{k,f}(t))\}$ and/or templated text referencing $P$.
\end{itemize}

Domain priors $\mathcal{D}$ (ontologies, invariants, constraints) and context $\mathcal{V}$ (subject/machine baselines) gate implausible combinations and shift thresholds. $\Theta$ parameterizes the mapping (from linear heads to compact MLPs with concept heads).

\paragraph{Auditability.}
For each $m_k$, SCI stores $\mathrm{TopFeat}(k,t)=\arg\max_f |a_{k,f}(t)|$ with signs, enabling deterministic rationales (“200 Hz line + sidebands ↑; sensor-4 temperature ↑”).

\subsection{Interpreter, markers, and clarity $SP_\theta$:}
\label{sec:sp-definition}

So far $SP(t)$ has been defined as a scalar quality signal over time windows. For the learning-theoretic analysis we now move to a per-example notation and write $x$ for a generic input window (e.g., $x = X_{t:t+\Delta t}$). A base model produces both a task output and an internal representation,
\[
f_\theta(x) = \bigl(y_\theta(x),\, h_\theta(x)\bigr), \qquad
h_\theta(x) \in \mathbb{R}^m,
\]
where $y_\theta(x)$ is the prediction (class probabilities or regression output) and $h_\theta(x)$ is a latent feature vector computed on top of the interpretable stack $P(t,s)$.

\paragraph{Marker head.}
SCI attaches a low-capacity \emph{marker head} $g_\theta$ to $h_\theta(x)$:
\[
P_\theta(x) = g_\theta\bigl(h_\theta(x)\bigr) \in \mathbb{R}^k,
\]
where $k$ is the number of \emph{cognitive markers}. In practice $g_\theta$ is a linear layer or shallow MLP (at most two layers), so that markers must reuse structure already present in $h_\theta(x)$ rather than learning an unconstrained auxiliary model. We convert marker logits to a probability vector
\[
q(x) = \operatorname{softmax}\bigl(P_\theta(x)\bigr), \qquad
q_i(x) = \frac{\exp(P_{\theta,i}(x))}
               {\sum_{j=1}^k \exp(P_{\theta,j}(x))},
\]
and define the Shannon entropy
$H(q(x)) = -\sum_{i=1}^k q_i(x)\,\log q_i(x)$.

\paragraph{Marker-based clarity.}
The SCI clarity signal is the normalized entropy of $q(x)$:
\begin{equation}
  SP_\theta(x)
  \;=\;
  1 - \frac{H\bigl(q(x)\bigr)}{\log k}
  \;\in\; [0,1].
  \label{eq:marker-clarity}
\end{equation}
Here $SP_\theta(x)\approx 1$ when a small number of markers dominate (low entropy; a “focused” internal state) and $SP_\theta(x)\approx 0$ when marker usage is diffuse (high entropy). Normalizing by $\log k$ makes $SP_\theta$ comparable across different numbers of markers $k$. In the streaming setting we can recover the original $SP(t)$ by aggregating per-window clarity, e.g.\ $SP(t) = \mathbb{E}_x[SP_\theta(x)]$ over windows ending at time $t$. We therefore treat $SP_\theta(x)$ as the per-example realization of the Surgical Precision signal.

\subsection{Closed-Loop Update and Lyapunov Energy}
\label{sec:controller}

Define $\Delta SP(t)=SP^\star(t)-SP(t)$ with time-varying target $SP^\star(t)\in(0,1]$ (policy/ethics/physics-calibrated; specified in \S3). SCI updates $\Theta$ in discrete time:
\[
\Theta_{t+1}=\mathrm{Proj}_{\mathcal{C}}\!\left[\Theta_t+\eta_t\big(\Delta SP(t)\,\nabla_{\Theta}SP(t)+\lambda_h\,u_h(t)\big)\right]\tag{1}
\]
\noindent\textbf{Projection operator.}
We use the Euclidean projection onto the feasible set $\mathcal{C}$:
\[
\mathrm{Proj}_{\mathcal{C}}(x) \;=\; \arg\min_{y\in\mathcal{C}} \|y-x\|_{2},
\]
implemented as coordinate-wise clipping for box constraints and an optional group-lasso proximal step to enforce structured sparsity. This guarantees $\Theta_{t+1}\in\mathcal{C}$ each update.
Here, $u_h(t)$ is a bounded human-correction signal derived from feedback on markers/rationales, $\lambda_h\ge 0$ is the human-gain, and $\eta_t>0$ is the step size.

\paragraph{Assumptions.}
\begin{itemize}
    \item (A1) $SP(t)=\phi(\Theta_t;P,\mathcal{D},\mathcal{V})$ is $L$-smooth in $\Theta$.
    \item (A2) $\|\nabla_{\Theta}SP(\Theta)\|\le G$ on $\mathcal{C}$.
    \item (A3) $w_f(t)$ vary slowly (bounded variation); measurement noise in $SP$ has bounded variance.
    \item (A4) $\|u_h(t)\|\le U$; $0\le\lambda_h\le\bar{\lambda}$.
    \item (A5) $SP^\star(t)$ is piecewise constant or Lipschitz (slow drift).
\end{itemize}

\paragraph{Lyapunov candidate and descent.}
Let $V(t)=\tfrac12(\Delta SP(t))^2$. A one-step expansion of (1) under (A1--A2) yields
\[
V(t+1)-V(t)\le -\eta_t\,\mu\,(\Delta SP(t))^2+\eta_t\,\lambda_h\,|\Delta SP(t)|\,\|u_h(t)\|+O(\eta_t^2L),
\]
for some $\mu>0$ depending on curvature of $SP$ along $\nabla_{\Theta}SP$. Using (A4) and Cauchy--Schwarz,
\[
V(t+1)-V(t)\le -\eta_t\big(\mu-\lambda_h U c\big)(\Delta SP(t))^2+O(\eta_t^2L),
\]
where $c$ upper-bounds the local sensitivity of $SP$ to $u_h$. Thus, with $\eta_t\le\eta_{\max}$ and human-gain budget $\lambda_h<\mu/(Uc)$, $V$ decreases monotonically up to $O(\eta_t^2)$ terms, implying $\Delta SP(t)\to 0$ or a small noise-induced neighborhood.

\paragraph{Safeguards (controller-agnostic).}
\begin{itemize}
    \item \textbf{Rollback:} if $SP$ decreases for $K$ consecutive steps, revert to the last checkpoint $\Theta^{\mathrm{ckpt}}$.
    \item \textbf{Trust region / projection:} enforce $\|\Theta_{t+1}-\Theta_t\|\le\rho$.
    \item \textbf{Gain scheduling:} decay $\lambda_h$ when user disagreement is high or rationale uncertainty is large.
    \item \textbf{Confidence gating:} apply large updates only when $|\Delta SP|$ exceeds a persistence threshold (EMA over a window).
\end{itemize}
These safeguards provide input-to-state stability of the closed loop even with noisy labels or sporadic human corrections.

\paragraph{Intuition.}
The condition $\dot V<0$ (discrete $V(t+1)-V(t)<0$) means explanation quality will not oscillate wildly: with a bounded human-gain budget $\lambda_h<\mu/(Uc)$, $V=\tfrac12(\Delta SP)^2$ decreases monotonically up to bounded noise, so $\Delta SP$ converges to zero or a small neighborhood. See Appendix~\ref{app:lyapunov-proof} for the full formal derivation and proof of the Lyapunov descent result.

\subsection{A Guided Walk-Through of Figure~\ref{fig:overview}}
\label{subsec:walkthrough-ecg-technical}
\textbf{Concrete example (ECG lead detachment).}
Consider an ICU ECG where a limb lead detaches at time $t_0$. In (M2), $\Pi$ decomposes the raw trace into rhythmic bands (e.g., 0.5--40\,Hz), low-frequency trend, and cross-sensor coherence. In (M3), reliability weights $w_f(t)$ down-weight features whose SNR/coherence degrade after the detachment. The interpreter $\psi_{\Theta}$ (M4) still proposes a marker (e.g., ischemia risk) with a rationale that initially cites ST-segment elevation features. The SP evaluator (M5) detects drops in $\kappa_1$ (clarity) and $\kappa_3$ (domain consistency), so $SP(t)$ falls and $\Delta SP(t)=SP^\star-SP(t)$ rises. When $|\Delta SP|>\gamma$, the controller (M6) applies the projected update $\Theta \leftarrow \mathrm{Proj}_{\mathcal{C}}[\Theta + \eta(\Delta SP \nabla_{\Theta} SP + \lambda_h u_h)]$. Within $3$--$5$ windows, the top features pivot from ischemia-like morphology to artifact-consistent bands and coherence loss; $SP(t)$ recovers.

\subsection{Lyapunov-style clarity objective and marker regularization}
\label{sec:lyapunov}

The control view of SCI is encoded directly into the training objective. Rather than hand-designing a dynamical update law in parameter space, we treat \emph{clarity misalignment} as a Lyapunov-style energy and minimize it jointly with the task loss under explicit anti-collapse regularizers.

\paragraph{Task-anchored target clarity.}
Let $x$ denote a generic input window (e.g.\ $x = X_{t:t+\Delta t}$) and $SP_\theta(x)$ the marker-based clarity defined in Eq.~\eqref{eq:marker-clarity}. We first construct a task-quality score $\tilde{R}(x;\theta)\in[0,1]$ from the current prediction $y_\theta(x)$:
\[
\tilde{R}(x;\theta) =
\begin{cases}
\text{margin}(x) = \max\bigl(0,\,p_c(x)-p_{(2)}(x)\bigr) & \text{(classification)},\\[4pt]
\exp\!\bigl(-\text{huber}(y_\theta(x)-y_{\text{true}})\bigr) & \text{(regression)},
\end{cases}
\]
where $p_c(x)$ and $p_{(2)}(x)$ are the top-1 and top-2 class probabilities and $\text{huber}(\cdot)$ is the Huber loss. We then apply a stop-gradient operator and a monotone link $\psi:[0,1]\to[0,1]$:
\[
R_\theta(x) = \operatorname{sg}\bigl(\tilde{R}(x;\theta)\bigr),
\qquad
SP^\star(x) = \psi\bigl(R_\theta(x)\bigr),
\]
with a default choice $\psi(r) = \sigma\bigl(\alpha(r-\beta)\bigr)$ for tunable slope $\alpha$ and midpoint $\beta$. The stop-gradient ensures that $SP^\star(x)$ is treated as a \emph{fixed} target for clarity and cannot be improved by trivially manipulating $y_\theta(x)$.

\paragraph{Interpretive error and Lyapunov energy.}
For each example we define the interpretive error
\[
\Delta SP_\theta(x) = SP^\star(x) - SP_\theta(x),
\]
and the Lyapunov-style energy
\begin{equation}
  V(\theta)
  \;=\;
  \mathbb{E}_{x\sim D}\bigl[(\Delta SP_\theta(x))^2\bigr],
  \label{eq:lyapunov-energy}
\end{equation}
which measures the expected misalignment between desired and actual clarity. Minimizing $V(\theta)$ encourages high clarity when the model is confident and correct, and low clarity (high entropy) when task quality is poor.

\paragraph{Marker health regularizers.}
To prevent degenerate solutions (e.g.\ global marker collapse or saturated $SP_\theta$), SCI augments $V(\theta)$ with a bundle of regularizers. Let $\bar{q} = \mathbb{E}_{x\sim D}[q(x)]$ denote the average marker distribution and let $\mathcal{U}_k$ be the uniform distribution on $k$ markers. We define:
\begin{align}
R_{\mathrm{div}}(\theta)
  &= \mathrm{KL}\bigl(\bar{q}\,\Vert\,\mathcal{U}_k\bigr)
    = \sum_{i=1}^k \bar{q}_i \log\bigl(k\,\bar{q}_i\bigr),
    &&\text{(diversity: avoids global collapse)}, \\
R_{\mathrm{band}}(\theta)
  &= \Bigl(\mathbb{E}_{x}[SP_\theta(x)] - \mu_{\mathrm{target}}\Bigr)^2,
    &&\text{(band constraint: avoids $SP_\theta\!\approx 0$ or $1$)}, \\
R_{\mathrm{stab}}(\theta)
  &= \mathbb{E}_{x\sim D,\;t\sim T}
     \Bigl[ \bigl(SP_\theta(t(x)) - SP_\theta(x)\bigr)^2 \Bigr],
    &&\text{(stability: robustness to task-natural transforms)}.
\end{align}
Here $T$ is a task-specific family of natural perturbations (e.g.\ small affine transformations in vision, sensor jitter in time series), and $\mu_{\mathrm{target}}\in(0,1)$ is a target mean clarity (typically in $[0.5,0.8]$) used to keep $SP_\theta$ numerically in a well-conditioned band. The combined marker health term is
\begin{equation}
  R_{\mathrm{marker}}(\theta)
  \;=\;
  \alpha_{\mathrm{div}} R_{\mathrm{div}}
  + \alpha_{\mathrm{band}} R_{\mathrm{band}}
  + \alpha_{\mathrm{stab}} R_{\mathrm{stab}},
  \label{eq:marker-regularizer}
\end{equation}
with nonnegative weights $\alpha_{\mathrm{div}},\alpha_{\mathrm{band}},\alpha_{\mathrm{stab}}$.

\paragraph{Total training objective.}
Putting these pieces together, SCI trains the interpreter by minimizing
\begin{equation}
  L_{\mathrm{total}}(\theta)
  \;=\;
  L_{\mathrm{task}}(\theta)
  \;+\;
  \lambda\,V(\theta)
  \;+\;
  \gamma\,R_{\mathrm{marker}}(\theta),
  \label{eq:total-loss}
\end{equation}
where $L_{\mathrm{task}}$ is the standard prediction loss (cross-entropy, MSE, etc.), $\lambda\ge 0$ controls the strength of clarity alignment, and $\gamma\ge 0$ controls the strength of the marker regularization bundle. In practice $L_{\mathrm{total}}$ is optimized by stochastic gradient methods over minibatches, and $\lambda$ is swept to trace a Pareto frontier between task performance and interpretive stability.

\paragraph{Interpretability as a stabilizable objective.}
The Lyapunov energy $V(\theta)$ plays the role of an interpretive potential: when $\lambda>0$, gradient descent on $L_{\mathrm{total}}$ drives $\Delta SP_\theta(x)$ toward zero on the data distribution, subject to the non-degeneracy enforced by $R_{\mathrm{marker}}(\theta)$. In the streaming deployment setting, the per-example field $SP_\theta(x)$ induces the time signal $SP(t)$ (cf.\ \S\ref{sec:sp-definition}), so that reductions in $V(\theta)$ correspond empirically to reductions in the observed interpretive error $\Delta SP(t) = SP^\star(t) - SP(t)$. We view this as a discrete-time, data-driven analog of Lyapunov stability: interpretability is no longer a static report, but a stabilizable state whose misalignment energy can be explicitly minimized.

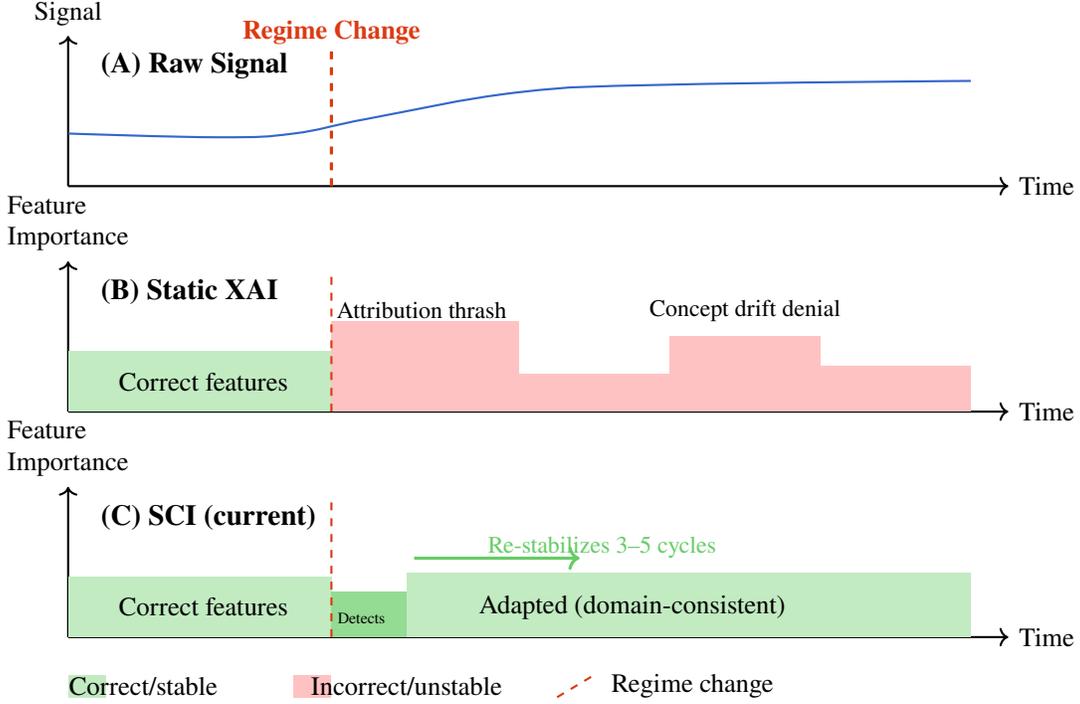
\begin{figure}[t]
\centering
\captionsetup{justification=centering,font=small}

\begin{adjustbox}{max width=\linewidth}
\begin{tikzpicture}[every node/.style={font=\small}]

\definecolor{signal}{RGB}{51,102,204}
\definecolor{event}{RGB}{220,57,18}
\definecolor{bad}{RGB}{255,102,102}
\definecolor{good}{RGB}{102,204,102}

\begin{scope}[shift={(0,6)}]
  \draw[thick,->] (0,0) -- (12.5,0) node[right]{Time};
  \draw[thick,->] (0,0) -- (0,2) node[above]{Signal};
  \draw[signal, thick] plot[smooth,tension=0.7] coordinates{
    (0,0.7) (2,0.65) (3,0.7) (4,0.9) (6,1.25) (8,1.35) (12,1.4)
  };
  \draw[event,dashed,very thick] (3.5,0) -- (3.5,1.8);
  \node[event] at (3.5,2.05) {\textbf{Regime Change}};
  \node[anchor=west,font=\bfseries] at (0.3,1.6) {(A) Raw Signal};
\end{scope}

\begin{scope}[shift={(0,3)}]
  \draw[thick,->] (0,0) -- (12.5,0) node[right]{Time};
  \draw[thick,->] (0,0) -- (0,2) node[above,align=left]{Feature\\Importance};
  \fill[good!40] (0,0)   rectangle (3.5,0.8);
  \fill[bad!40]  (3.5,0) rectangle (6,1.2);
  \fill[bad!40]  (6,0)   rectangle (8,0.5);
  \fill[bad!40]  (8,0)   rectangle (10,1.0);
  \fill[bad!40]  (10,0)  rectangle (12,0.6);

  \node at (1.8,0.4) {Correct features};
  \node at (4.7,1.35) {\footnotesize Attribution thrash};
  \node at (9,1.35)   {\footnotesize Concept drift denial};

  \draw[event,dashed,thick] (3.5,0) -- (3.5,1.8);
  \node[anchor=west,font=\bfseries] at (0.3,1.6) {(B) Static XAI};
\end{scope}

\begin{scope}[shift={(0,0)}]
  \draw[thick,->] (0,0) -- (12.5,0) node[right]{Time};
  \draw[thick,->] (0,0) -- (0,2) node[above,align=left]{Feature\\Importance};
  \fill[good!40] (0,0)   rectangle (3.5,0.8);
  \fill[good!70] (3.5,0) rectangle (4.5,0.6);
  \fill[good!40] (4.5,0) rectangle (12,0.85);

  \node at (1.8,0.4) {Correct features};
  \node[font=\tiny] at (3.9,0.25) {Detects};
  \node at (7.5,0.4) {Adapted (domain-consistent)};

  \draw[good,->,very thick] (4.6,1.05) -- (6.8,1.05);
  \node[good] at (7.1,1.2) {\footnotesize Re-stabilizes 3--5 cycles};

  \draw[event,dashed,thick] (3.5,0) -- (3.5,1.8);
  \node[anchor=west,font=\bfseries] at (0.3,1.6) {(C) SCI (current)};
\end{scope}

\begin{scope}[shift={(0,-0.8)}]
  \fill[good!40] (0,0) rectangle (0.5,0.3);
  \node at (1,0.15) {Correct/stable};

  \fill[bad!40] (3,0) rectangle (3.5,0.3);
  \node at (4.5,0.15) {Incorrect/unstable};

  \draw[event,dashed,thick] (6.5,0) -- (7,0.3);
  \node at (8.3,0.15) {Regime change};
\end{scope}

\end{tikzpicture}
\end{adjustbox}

\caption{\textbf{Static XAI vs.\ SCI under a regime change.}
Colored bullets below each panel show $\kappa=[\kappa_1,\kappa_2,\kappa_3,\kappa_4]$, with the component(s) that trigger adaptation highlighted when $|\Delta SP|>\gamma$. Dashed vertical lines mark regime-change events prompting Eq.~(1) updates. Markers use distinct shapes and colors so they remain distinguishable under common color-vision deficiencies and in grayscale.}
\label{fig:failure_timeline}
\end{figure}

\begin{figure}[t]
\centering
\captionsetup{justification=centering}
\begin{adjustbox}{max width=\linewidth}
\begin{tikzpicture}[
  every node/.style={font=\small},
  box/.style={draw, rounded corners, align=center, inner sep=4pt, minimum width=2.8cm, minimum height=1cm},
  comp/.style={draw, rounded corners, align=center, inner sep=3pt, minimum width=2.4cm, minimum height=0.9cm},
  arr/.style={-Latex, line width=0.8pt}
]

\node[font=\bfseries] at (0,6.2) {Feature Bank $P(t,s)$};
\node[box, fill=gray!20, minimum width=6.8cm] (raw) at (0,5) {Raw Multimodal Signals\\$X(t)\in\mathbb{R}^m$, $s\in\mathcal{S}$};
\node[box, fill=orange!20] (pi) at (0,3.8) {$\Pi$\\Decomposition};
\draw[arr, blue!70] (raw) -- (pi);
\node[comp, fill=green!10] (R)  at (-4,2.2) {$R(t)$\\Rhythmic};
\node[comp, fill=green!10] (T)  at (-1.3,2.2) {$T(t)$\\Trend};
\node[comp, fill=green!10] (S)  at (1.3,2.2) {$S(s)$\\Spatial};
\node[comp, fill=green!10] (C)  at (4,2.2) {$C(t,s)$\\Cross-modal};
\node[comp, fill=purple!10, minimum width=3.2cm] (Pstar) at (0,1.0) {$\Pi^{\star}(t,s)$\\Compact (PCA/AE)};

\draw[arr, blue!70] (pi) -- (R);
\draw[arr, blue!70] (pi) -- (T);
\draw[arr, blue!70] (pi) -- (S);
\draw[arr, blue!70] (pi) -- (C);
\draw[arr, blue!70] (pi) -- (Pstar);

\node[
  draw, rounded corners, fill=yellow!10,
  font=\scriptsize,
  minimum width=8cm,
  align=center
] at (0,-0.7) {
  Reliability:\quad
  $w_f(t)=\frac{e^{\gamma z_f(t)}}{\sum_g e^{\gamma z_g(t)}}$,\;
  $z_f=\log\mathrm{SNR}_f+\alpha\,\mathrm{Pers}_f+\beta\,\mathrm{Coh}_f$
};

\begin{scope}[xshift=10cm]
\node[font=\bfseries] at (0,6.2) {Surgical Precision $SP(t)$};
\node[box, fill=red!10, minimum width=6.5cm] (sp) at (0,5) {
  $SP(t)=w^\top\kappa(t)$\\
  {\footnotesize $\kappa=[\kappa_1,\kappa_2,\kappa_3,\kappa_4]^\top \in[0,1]^4$}
};
\node[comp, fill=cyan!20, minimum width=6cm] (k1) at (0,3.8) {
  $\kappa_1$ Clarity\\{\footnotesize Selectivity, SNR}
};
\node[comp, fill=cyan!20, minimum width=6cm] (k2) at (0,2.6) {
  $\kappa_2$ Pattern Strength\\{\footnotesize Amplitude, stability}
};
\node[comp, fill=cyan!20, minimum width=6cm] (k3) at (0,1.4) {
  $\kappa_3$ Domain Consistency\\{\footnotesize Physics/ontology constraints}
};
\node[comp, fill=cyan!20, minimum width=6cm] (k4) at (0,0.2) {
  $\kappa_4$ Predictive Alignment\\{\footnotesize Rolling AUC/F1}
};

\draw[arr, blue!70] (sp) -- (k1);
\draw[arr, blue!70] (k1) -- (k2);
\draw[arr, blue!70] (k2) -- (k3);
\draw[arr, blue!70] (k3) -- (k4);
\node[
  draw, rounded corners, fill=yellow!10,
  font=\scriptsize,
  minimum width=6.3cm,
  align=center
] at (0,-1.0) {
  Each $\kappa_i$ mapped to $[0,1]$\\
  via monotone calibrators $\sigma_i$
};
\end{scope}

\draw[blue!70, dashed, -Latex, line width=0.9pt] (4.5,3.2) -- ++(1.2,0)
  node[midway, above, font=\small] {Feeds $\psi_\Theta$};
\end{tikzpicture}
\end{adjustbox}

\caption{\textbf{Decomposition and SP anatomy.} Left: $\Pi$ produces semantic components used in $P(t,s)$. Right: $SP(t)$ aggregates calibrated components $\kappa_1$–$\kappa_4$ via convex weights $w \in \Delta^3$. All $\kappa$ components are encoded with distinct colors and marker shapes (and line styles) so they remain distinguishable under common color-vision deficiencies and in grayscale.}
\label{fig:decomp_sp}
\end{figure}
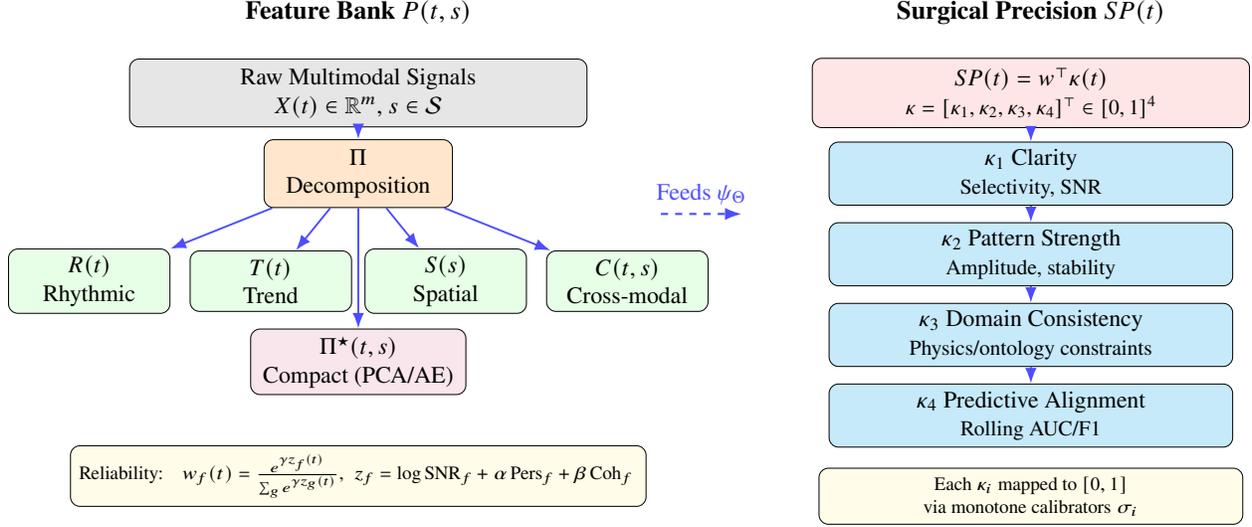

\subsection{Practical Estimation and Identifiability:}
\paragraph{Gradients.}
When $\psi_{\Theta}$ is differentiable end-to-end, $\nabla_{\Theta}SP$ is obtained by automatic differentiation. For symbolic or rule components, use finite-difference or implicit differentiation with straight-through estimators on $\sigma_i$.
\paragraph{Calibration.}
The $\sigma_i$ calibrators are monotone and learned on validation data (isotonic/logistic), preserving order and keeping $SP\in[0,1]$.
\paragraph{Identifiability.}
Any two interpretations $I_1,I_2$ that induce the same $\kappa$ yield equal $SP$. Distinct $\kappa$ map to distinct $SP$ provided $w\in\Delta^3$ has nonzero entries; thus $SP$ is order-identifiable in $\kappa$.
\paragraph{Drift robustness.}
Reliability weights $w_f(t)$ update via EMA with bounded rates; if sensor-health flags drop below a threshold, affected features are masked ($w_f\to0$), preventing explanation drift.
\paragraph{Windowing.}
$\kappa_4$ (predictive alignment) may use lagged outcomes; overlapping windows or exponential decay integrate delayed feedback without destabilizing the fast loop on $\kappa_{1:3}$.
\paragraph{Bridge to §6 (Architecture).} Section 6 instantiates $\Pi$, $\psi_{\Theta}$, and update (1) with concrete modules (decomposition filters, concept heads, calibrators), describes gradient pathways for $SP$, and implements the safeguards (rollback, trust-region, gain scheduling) used in our experiments.

\section{Technical Architecture}
\begin{table}[t]
\centering
\small
\begin{tabular}{@{}llp{0.55\linewidth}@{}}
\toprule
\textbf{Module} & \textbf{Role} & \textbf{Key I/O} \\
\midrule
(M1) Ingestion & Stream, validate, window & $X(t)\!\in\!\mathbb{R}^m \to X_{t:t+\Delta t}$, health flags $q$ \\
(M2) Decomposition $\Pi$ & Featureize into named blocks & $X_{t:t+\Delta t} \to \{R,T,S,C,\Pi^\star\}$ \\
(M3) Composer & Assemble $P(t,s)$ \& reliability & blocks $+$ $q \to P(t,s)$, weights $\{w_f(t)\}$, masked $P_w$ \\
(M4) Interpreter $\psi_\Theta$ & Markers, confidences, rationales & $P_w, \mathcal{D}, \mathcal{V} \to I(t)=\{(m_k,p_k,r_k)\}$ \\
(M5) SP Evaluator & Compute $SP(t)$, $\kappa_{1:4}$ & $(P_w,I,\mathcal{D},\mathcal{V}) \to SP(t)=w^\top\kappa(t)$ \\
(M6) Controller & Update $\Theta$ via $\Delta SP$ 
& $SP,\; SP^\star,\; \nabla_{\Theta} SP,\; u_h \to \Theta_{t+1}$ \\
(M7) UI/Buffer & Visualize \& collect feedback & $(I,SP)$, user events $\to \mathcal{B}$ \\
\bottomrule
\end{tabular}
\caption{SCI modules at a glance. Roles and data flow align with \S6.1.}
\label{tab:modules}
\end{table}

SCI is a closed-loop, multi-module system that ingests raw signals, constructs the interpretable state $P(t,s)$, produces $I(t)$, evaluates $SP(t)$, and adapts $\Theta$ online via $\Delta SP$. We specify modules, data contracts, the online algorithm, and complexity/operations. (Notation: $\Delta SP = SP^\star - SP$, consistent with §5.)

\subsection{System Modules and Data Contracts:}
\paragraph{(M1) Ingestion Layer\\.}
 Role: stream, validate, synchronize, and window raw multimodal signals.\\
 Input: $X(t)\in\mathbb{R}^{m}$, optional spatial index $s\in\mathcal{S}$.\\
 Output: aligned batch $X_{t:t+\Delta t}$, health flags $q$ (dropout, range checks).\\
 Invariants: monotone timestamps; fixed sampling metadata per window; basic imputation if $q$ flags minor gaps.

\paragraph{(M2) Decomposition Bank ($\Pi$)\\.}
 Role: convert $X$ to named multi-scale features.\\
 Input: $X_{t:t+\Delta t}$.\\
 Output: $\{R(t), T(t), S(s), C(t,s), \Pi^{\star}(t,s)\}$.\\
 Typical ops: FFT/STFT (spectral), CWT (time--frequency), EMD/VMD (adaptive modes), SSA (trend), coherence / cross-correlation / Granger (interactions), wavelet denoising.\\
 Invariant: every feature is typed and auditable (name, units, window).

\paragraph{(M3) Feature Composer (reliability-weighted $P$)\\.}
 Role: assemble $P(t,s)$ and compute reliability.\\
 Input: blocks from (M2) + health flags $q$.\\
 Output: $P(t,s)=[R,T,S,C,\Pi^{\star}]$; weights $\{w_f(t)\}$; masked view $P_w(t,s)$.\\
 Reliability: for each feature $f$,
\[
z_f(t)=\log \mathrm{SNR}_f + \alpha\,\mathrm{Pers}_f + \beta\,\mathrm{Coh}_f,\qquad w_f(t)=\frac{e^{\gamma z_f(t)}}{\sum_g e^{\gamma z_g(t)}},\ \sum_f w_f=1.
\]
EMA smoothing and bounded rate-of-change prevent thrash; failed-health features are down-weighted or omitted ($w_f\to0$).\\
 Invariant: $P$ is named and reliability-weighted before interpretation.

\paragraph{(M4) Knowledge-Guided Interpreter $\psi_{\Theta}$\\.}
 Role: map $P_w$ to interpretable output.\\
 Input: $P_w(t,s)$; priors $\mathcal{D}$ (ontologies, invariants); context $\mathcal{V}$ (subject/machine baselines).\\
 Output: $I(t)=\{(m_k,p_k(t),r_k(t))\}_{k=1}^{K}$.\\
 Implementation: lightweight heads (linear/MLP) plus concept heads constrained by $\mathcal{D}$; rationales record sparse attributions $\{(f,a_{k,f})\}$ and TopFeat lists for audit.\\
 Invariant: ontology/physics constraints gate implausible combinations before scoring.

\paragraph{(M5) SP Evaluator\\.}
 Role: compute $SP(t)\in[0,1]$ and components $\kappa_{1:4}$.\\
 Input: $(P_w,I(t),\mathcal{D},\mathcal{V})$ with optional lagged outcomes for $\kappa_4$.\\
 Output: $SP(t)=w^{\top}\kappa(t)$, $\kappa\in[0,1]^4$.\\
 Calibration: isotonic calibrators $\sigma_i$ by default; logistic when data are sample-limited.\\
 Logging: each step logs $(\kappa(t),w,\mathrm{TopFeat}(k,t))$ with hashes of inputs and parameters for deterministic audits.\\
 Invariant: rolling windows for stability; component logs support QA.

\paragraph{(M6) Adaptive Controller\\.}
 Role: update $\Theta$ when interpretation lags target.\\
 Input: $SP(t)$, target $SP^\star$, $\nabla_{\Theta}SP(t)$ (or finite-difference/STE), human signal $u_h(t)$ from buffer $\mathcal{B}$.\\
 Update:
\[
\Theta_{t+1}=\mathrm{Proj}_{\mathcal{C}}\!\left[\Theta_t+\eta_t\big(\Delta SP(t)\nabla_{\Theta}SP(t)+\lambda_h\,u_h(t)\big)\right],\quad \Delta SP=SP^\star-SP.
\]
\textbf{Human signal:} surrogate gradient from corrections--cross-entropy on corrected markers $m_k$ plus a hinge/contrastive term on rationale attributions $a_{k,f}$.\\
 Safeguards: threshold $\gamma$ (no-op zone), rollback on $K$ consecutive drops in $SP$, trust region $\|\Theta_{t+1}-\Theta_t\|\le\rho$, gain scheduling for $\lambda_h$, and runtime enforcement of $\lambda_h<\mu/(Uc)$ (per §5).\\
 Invariant: updates are monotone on $V=\tfrac12(\Delta SP)^2$ under §5 assumptions.

\paragraph{(M7) UI and Feedback Buffer\\.}
 Role: visualize $I(t)$, $SP(t)$; collect corrections.\\
 Input: latest $I$, $SP$; user actions (confirm/deny markers, rationale nudges, severity weights).\\
 Output: buffer $\mathcal{B}$ of structured feedback events (versioned to $\Theta$) and optional $\lambda_h$ hints.\\
 Invariant: all feedback is timestamped and scoped to the viewed data slice to avoid staleness.

\subsection{Online Algorithm (pseudocode):}
\begin{verbatim}
Initialize Theta <- Theta_0, checkpoints<-{Theta_0}, lambda_h <- lambda_h,0
repeat for each window [t, t + dt):
    (M1) INGEST
    X_batch, q <- ingest()
    (M2) DECOMPOSE
    R, T, S, C, Pi* <- Pi(X_batch)
    (M3) COMPOSE + RELIABILITY
    P, w <- compose_and_weight(R,T,S,C,Pi*, q)  # EMA, masking, softmax weights
    (M4) INTERPRET
    I <- psi_Theta(P, D, V)                     # {(m_k, p_k, r_k)}
    (M5) EVALUATE + LOG
    kappa <- components(P, I, D, V)             # kappa_1..4 in [0,1]
    SP <- w_kappa * kappa                       # convex weights
    log_step(kappa, w, TopFeat(I), hash(X_batch, Theta))
    dSP <- SP* - SP
    (M6) ADAPT
    if |dSP| > gamma:
        g <- grad_SP(Theta; P, I, kappa)        # autodiff or finite diff/STE
        u_h <- human_signal(B)                  # CE on m_k + hinge on a_k,f
        Theta_cand <- proj_C(Theta + eta * (dSP * g + lambda_h * u_h))
        if SP(Theta_cand) >= SP:
            Theta <- Theta_cand
            checkpoints.push(Theta)
        else:
            bad_updates <- bad_updates + 1
            if bad_updates >= K:
                Theta <- checkpoints.last()
                bad_updates <- 0
    else:
         log_state()
    (M7) UI + BUFFER
    visualize(I, SP); B <- B + user_feedback()
    
    # periodic meta-update (slow path)
    if t mod T_meta == 0:
        meta_update(Theta, D, V, w_kappa; B)
        B <- empty
\end{verbatim}

Defaults: $\gamma$ = 1--2 $\times$ MAD of recent $SP$; $K\in\{2,3\}$;
trust region $\rho$ chosen to keep $SP$ non-decreasing on a one-step holdout;
$T_{\text{meta}}$ keeps slow meta-updates from perturbing the fast loop.

\subsection{Computational Complexity and Deployment Profile:}

\paragraph{Concrete latencies (reference pipeline).}
On our reference pipeline, controller updates add $\sim 12\!\pm\!2$\,ms (GPU) per window; total end-to-end latency is $\sim 79\!\pm\!8$\,ms at $n\!\approx\!100$ features. Scaling to $n\!=\!500$ yields $\sim 312$\,ms with block-sparse interactions ($k\!=\!20$), maintaining real-time throughput at $\sim$12\,Hz.

\paragraph{Memory and energy profile.}
Memory overhead is dominated by the interaction buffers and $\Pi$’s feature bank; the controller maintains negligible state. On GPU edge devices (10--15\,W class), we sustain $\sim$12\,Hz at $n\!\approx\!100$ features; CPU-only pipelines remain sub-second at moderate $n$ with caching and block-sparse interactions.

\paragraph{Per-window costs (dominant terms).}
Decomposition (M2): FFT/STFT $O(N\log N)$; CWT $O(NJ)$ for $J$ scales; EMD/VMD $O(JN\cdot \text{iter})$.\\
Interactions (part of $C$): naïve coherence/correlation across $n$ features $O(n^2)$.\\
Mitigation: block-diagonal by modality (vision, vibration, EEG, etc.), with $k$-NN sparsification within blocks and sketching for long tails.\\
Interpreter (M4): small MLP/linear heads $O(d)$--$O(dh)$; autodiff adds a constant factor.\\
SP (M5): $O(|P|)$ to accumulate $\kappa_{1:4}$ (rolling stats); lagged $\kappa_4$ uses incremental counters.\\
Update (M6): one projected step $O(d)$ (box constraints) to $O(d\log d)$ (sparsity projections).

\paragraph{Real-time viability.}
Moderate $n$ (hundreds of features): CPU pipeline with vectorized FFT and cached coherences $\Rightarrow$ sub-second latency.\\
Large $n$ (thousands+): GPU-offload $\Pi$; pin (M2--M3) and (M4--M6) to separate executors (producer--consumer); mini-batch interactions.

\paragraph{Parallelization and caching.}
Pipelining: (M2) and (M5) overlap; (M4) waits only on $P_w$.\\
Caching: rolling means/variances for $\kappa$; memoize band powers and coherence on overlapping windows.\\
Asynchrony guard: feedback events in $\mathcal{B}$ are versioned to $\Theta$; the controller ignores stale entries.

\paragraph{Safety and robustness.}
Rollback and trust-region in the controller; human-gain budget enforced at runtime.\\
Drift response: if feature health declines, $w_f \to 0$ and $\kappa_3$ penalizes implausible interpretations; the controller tempers $\eta_t/\lambda_h$ until $SP$ stabilizes.\\
Audit: each $I(t)$ carries TopFeat plus $SP$ component logs $(\kappa, w)$ to support failure analysis.

\subsection{Interfaces (concise API):}
$\Pi$: features = decompose(X\_batch, cfg) $\to$ \{R,T,S,C,$\Pi*$\} with metadata.\\
 Composer: P\_w, weights = compose\_weight(features, health, ema\_state)\\
 Interpreter: I = interpret(P\_w, D, V, $\Theta$) (see §5.2)\\
 SP: SP, $\kappa$ = evaluate\_sp(P\_w, I, D, V, calibs) (default: isotonic; logistic if sample-limited; see §5.3)\\
 Controller: $\Theta'$ = adapt($\Theta$, SP, $SP^\star$, grad\_sp, $u_h$, $\lambda_h$) ($u_h$: CE on $m_k$ + hinge on $a_{k,f}$; see §5.4)\\
 UI/Buffer: B = collect\_feedback(events, $\Theta_{\mathrm{version}}$) (see §6.1)\\
 All functions are pure with respect to inputs (except controller checkpoints), enabling deterministic replay.

\paragraph{Bridge to §7 (Experiments).} Section 7 reports latency and throughput, ablations for reliability weighting and interaction sparsification, stability curves for $V=\tfrac12(\Delta SP)^2$, and task metrics (AUC and F1) under controlled drift.

\paragraph{Minimal runnable loop (for practitioners).}
The following 12-line loop implements SCI's core control principle without external dependencies:
\begin{verbatim}
# Pseudocode: SCI online control loop
for X_batch in stream():
    P, w      = decompose_and_weight(X_batch)         # M2–M3
    I         = psi_theta(P, D, V)                    # M4
    SP, kappa = evaluate_SP(P, I, D, V)               # M5
    dSP       = SP_star - SP
    if abs(dSP) > gamma:                              # no-op zone
        g    = grad_SP(theta, P, I, kappa)            # autodiff or STE
        u_h  = human_signal(buffer)                   # optional
        step = eta * (dSP * g + lambda_h * u_h)
        theta_candidate = project(theta + step)       # box / group constraints
        if evaluate_SP(P, psi_theta(P, D, V, theta_candidate), D, V) >= SP:
            theta = theta_candidate                   # monotone safeguard
\end{verbatim}
This snippet embodies SCI's central idea: \emph{treat $SP$ as a regulated signal and adapt $\Theta$
only when interpretive error $\Delta SP$ exceeds a persistence threshold}.

\section{Evaluation}
\label{sec:evaluation}

In this first empirical study we evaluate a minimal SCI instantiation: $\Pi$ collapses to the latent features of a standard dropout network; $SP$ reduces to normalized predictive entropy; and the controller uses a threshold policy on $\Delta SP$. This deliberately underuses the full architecture of Sections 5--6; our goal is to test whether even this reduced SCI layer already exhibits metacognitive allocation and usable safety signals. Richer decompositions and human-feedback loops are left for future work. Our goal in this section is not to claim state-of-the-art performance, but to probe whether SCI behaves as a sensible metacognitive controller around ordinary stochastic classifiers. Concretely, we ask:

\begin{enumerate}
    \item Does SCI allocate more computation to examples that it ultimately misclassifies than to those it gets right?
    \item Does the interpretive error $\Delta SP$ act as a useful safety signal for detecting errors?
    \item Can SCI trade computation for accuracy more efficiently than a blind fixed-budget ensemble?
    \item Where does the mechanism break down (e.g., under extreme distribution shift)?
\end{enumerate}

Throughout, we treat SCI as a light-weight wrapper around a fixed predictive model with Monte–Carlo dropout, and instantiate the interpretive state $SP(t)$ as an entropy-based ``Surgical Precision'' signal (see \S\ref{sec:sp-definition}). For each experiment we log task error, average inference steps, and safety metrics derived from $\Delta SP$.

\subsection{Tasks and models}

We evaluate on three representative domains that capture vision, medical, and industrial monitoring:

\begin{itemize}
    \item \textbf{MNIST digits (vision).} A small CNN with dropout in the penultimate layer is trained on a standard MNIST split. At inference time the network is run in stochastic mode (dropout enabled) to support Monte--Carlo sampling.
    \item \textbf{MIT--BIH arrhythmia (medical ECG).} We follow a conventional binary framing (normal vs.\ arrhythmia) on MIT--BIH RR-interval / ECG streams, using 12,000 training and 2,000 test segments. The base model is a 1D convolutional classifier with dropout.
    \item \textbf{Rolling bearings (industrial vibration).} Short vibration windows from a standard run-to-failure bearing dataset are classified as healthy vs.\ fault using a 1D CNN with dropout. This domain is intentionally simple but representative of industrial condition monitoring.
\end{itemize}

In all cases the predictive architecture and training recipe are fixed; SCI only modifies how many stochastic forward passes are taken per input and how these are interpreted.

\subsection{Metrics and protocol}

For a given test input $x$ we obtain a sequence of predictive distributions $\{p_t(y \mid x)\}_{t=1}^T$ by repeatedly sampling the network with dropout enabled. We define
\begin{equation}
    SP(t) = 1 - \frac{H(p_t)}{\log K},
\end{equation}
where $H(\cdot)$ is the Shannon entropy and $K$ the number of classes, and choose a target $SP^\star$. The interpretive error is
\begin{equation}
    \Delta SP(t) = \bigl|SP^\star - SP(t)\bigr|.
\end{equation}

The SCI controller monitors $SP(t)$ and either stops (emitting the current predictive mean), continues sampling, or abstains when a budget $T_{\max}$ is exceeded. Unless otherwise specified we report averages over three random seeds and log:

\begin{itemize}
    \item \textbf{Task performance:} classification error rate on the test set.
    \item \textbf{Metacognitive allocation:} mean number of inference steps for correctly vs.\ incorrectly classified samples, and the full distributions of step counts.
    \item \textbf{Safety:} $\mathrm{AUROC}_{\Delta SP}$, the AUROC of $\Delta SP$ as a detector of errors; in some experiments we also compare against standard confidence.
    \item \textbf{Risk--coverage curves (MIT--BIH):} accuracy as a function of the fraction of samples retained when we reject high-$\Delta SP$ cases.
    \item \textbf{Compute--accuracy trade-offs (MIT--BIH):} accuracy vs.\ average number of Monte--Carlo samples, comparing SCI to fixed-size ensembles.
\end{itemize}

\subsection{SCI architecture in practice}

\begin{figure}[t]
\centering
\begin{tikzpicture}[
  scale=0.95,
  module/.style={draw, rectangle, rounded corners, minimum width=3.2cm, minimum height=1.3cm, align=center, font=\small, fill=blue!10},
  moduleG/.style={draw, rectangle, rounded corners, minimum width=3.2cm, minimum height=1.3cm, align=center, font=\small, fill=green!10},
  moduleC/.style={draw, rectangle, rounded corners, minimum width=3.2cm, minimum height=1.3cm, align=center, font=\small, fill=cyan!10},
  moduleO/.style={draw, rectangle, rounded corners, minimum width=3.2cm, minimum height=1.3cm, align=center, font=\small, fill=orange!15},
  moduleV/.style={draw, rectangle, rounded corners, minimum width=3.2cm, minimum height=1.3cm, align=center, font=\small, fill=violet!10},
  io/.style={font=\scriptsize, align=center},
  arrow/.style={-Latex, thick, blue!60},
  feedback/.style={-Latex, thick, orange!80, dashed}
]

\node[module] (m1) at (0,0) {\textbf{M1}\\Ingestion};
\node[module] (m2) at (4,0) {\textbf{M2}\\Decomposition $\Pi$};
\node[module] (m3) at (8,0) {\textbf{M3}\\Composer};
\node[moduleG] (m4) at (12,0) {\textbf{M4}\\Interpreter $\psi_\Theta$};
\node[moduleC] (m5) at (12,-3) {\textbf{M5}\\SP Evaluator};
\node[moduleV] (m7) at (12,-6) {\textbf{M7}\\UI/Buffer};
\node[moduleO] (m6) at (6,-3.5) {\textbf{M6}\\Controller};

\draw[arrow] (m1) -- (m2);
\draw[arrow] (m2) -- (m3);
\draw[arrow] (m3) -- (m4);
\draw[arrow] (m4) -- (m5);

\draw[arrow] (m3.south) |- (m5.west);
\draw[arrow] (m5.west) -- (m6.east);

\draw[arrow, dashed] (m4.south) |- (m7.north);
\draw[arrow] (m5) -- (m7);

\draw[feedback, very thick] (m6.north) |- ++(0,1.5) -| (m4.south west);
\draw[feedback] (m7.west) -| (m6.south);


\node[draw, rounded corners, fill=yellow!15, text width=3.2cm, font=\tiny, anchor=south] 
at (8,1.8) {
\textbf{Reliability weighting}\\
$w_f=\frac{e^{\gamma z_f}}{\sum_g e^{\gamma z_g}}$\\
$z_f=\log\mathrm{SNR}+\alpha \mathrm{Pers}+\beta\mathrm{Coh}$
};
\node[draw, rounded corners, fill=cyan!10, text width=3.2cm, font=\tiny, anchor=west]
at (14.0,-3.2) {
  \textbf{SP = $w^\top\kappa$}\\
  $\kappa_1$ clarity, $\kappa_2$ strength\\
  $\kappa_3$ consistency, $\kappa_4$ alignment
};
\node[draw, rounded corners, fill=green!10, text width=3.4cm, font=\tiny, anchor=north]
at (6,-5.5) {
\textbf{Safeguards}\\
Lyapunov $V=\tfrac12(\Delta SP)^2$\\
Rollback on $K$ drops\\
Trust region $\rho$\\
Gain budget $\lambda_h$\\
No-op zone $|\Delta SP|<\gamma$
};
\node[font=\bfseries\large, anchor=south west] at (-0.5,1.6) {SCI Module Pipeline (M1--M7)};

\end{tikzpicture}

\caption{SCI module pipeline used in our experiments. The base predictor and feature decomposition (M1--M3) are standard; SCI adds an interpreter, SP evaluator, controller, and buffer (M4--M7). In the present work, $SP$ is instantiated via entropy, but the structural decomposition into reliability-weighted features and safeguards remains applicable to richer instantiations.}
\label{fig:module_pipeline}
\end{figure}
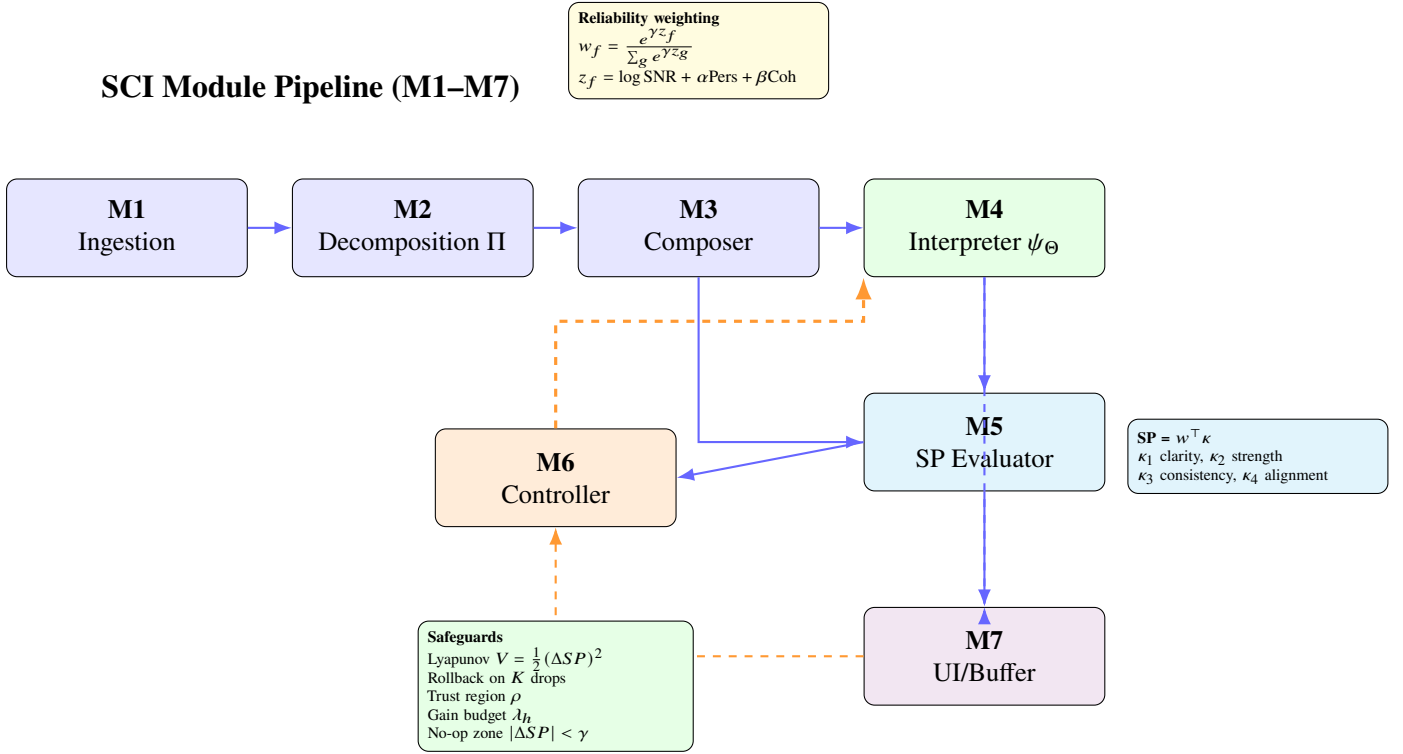

Figure~\ref{fig:module_pipeline} recalls the SCI module pipeline used in these experiments. The base predictor and feature extraction stack (M1–M3) are standard; SCI adds an interpreter $\psi_\Theta$ (M4), an SP evaluator (M5), a controller (M6), and a UI/buffer layer (M7). For the present empirical study, $SP$ is instantiated as the normalized entropy above, and the controller implements a simple gain-scheduled thresholding rule rather than the full general-purpose design from \S\ref{sec:controller}.

\subsection{Metacognitive allocation of computation}

\begin{figure*}[t]
    \centering
    \begin{tikzpicture}
      \begin{axis}[
          width=0.45\textwidth, height=5cm,
          title={MNIST: Steps (Green=Correct, Red=Incorrect)},
          xlabel={Inference Steps}, ylabel={Density},
          ybar, bar width=6pt,
          symbolic x coords={2,4,6,8,10,12,14,16},
          xtick=data,
          ymin=0, ymax=0.8,
          ymajorgrids=true,
          grid style={dashed, gray!30},
          legend pos=north east,
          legend style={nodes={scale=0.8, transform shape}},
          axis line style={gray},
          tick align=outside
      ]
      \addplot[fill=green!60!black, opacity=0.8, draw=none] coordinates {(2,0.7) (4,0.1) (6,0.05) (8,0.02) (10,0.01) (12,0.01) (14,0.01) (16,0)};
      \addplot[fill=red!60!black, opacity=0.8, draw=none] coordinates {(2,0.05) (4,0.05) (6,0.05) (8,0.05) (10,0.05) (12,0.05) (14,0.1) (16,0.6)};
      \legend{Correct, Incorrect}
      \end{axis}
    \end{tikzpicture}
    \hfill
    \begin{tikzpicture}
      \begin{axis}[
          width=0.45\textwidth, height=5cm,
          title={MIT-BIH: Steps (Green=Correct, Red=Incorrect)},
          xlabel={Inference Steps}, ylabel={Density},
          ybar, bar width=6pt,
          ymin=0, ymax=0.18,
          ymajorgrids=true,
          grid style={dashed, gray!30},
          legend pos=north west,
          legend style={nodes={scale=0.8, transform shape}},
          axis line style={gray},
          tick align=outside
      ]
      \addplot[fill=green!60!black, opacity=0.8, draw=none] coordinates {(5,0.1) (10,0.12) (15,0.08) (20,0.05) (25,0.02)};
      \addplot[fill=red!60!black, opacity=0.8, draw=none] coordinates {(5,0.02) (10,0.05) (15,0.1) (20,0.12) (25,0.15)};
      \legend{Correct, Incorrect}
      \end{axis}
    \end{tikzpicture}
    
    \caption{\textbf{SCI as a metacognitive controller.}
    Left: MNIST digit classification. Right: MIT--BIH arrhythmia detection.
    For each domain we plot the empirical distribution of inference steps under the SCI controller, decomposed into samples that are ultimately classified correctly (green) and incorrectly (red). On MNIST, errors receive roughly $3.6\times$ more steps than correct predictions (mean $2.84$ vs.\ $10.31$), while on MIT--BIH errors receive about $1.4\times$ more steps (mean $14.24$ vs.\ $19.86$). On the bearings dataset (not shown), SCI spends $5.83$ steps on correct windows and $22.14$ on the rare mistakes ($\approx 3.8\times$ more). These patterns indicate that SCI reallocates computational budget as a function of difficulty rather than applying a fixed inference graph to all inputs.}
    \label{fig:key_results}
\end{figure*}
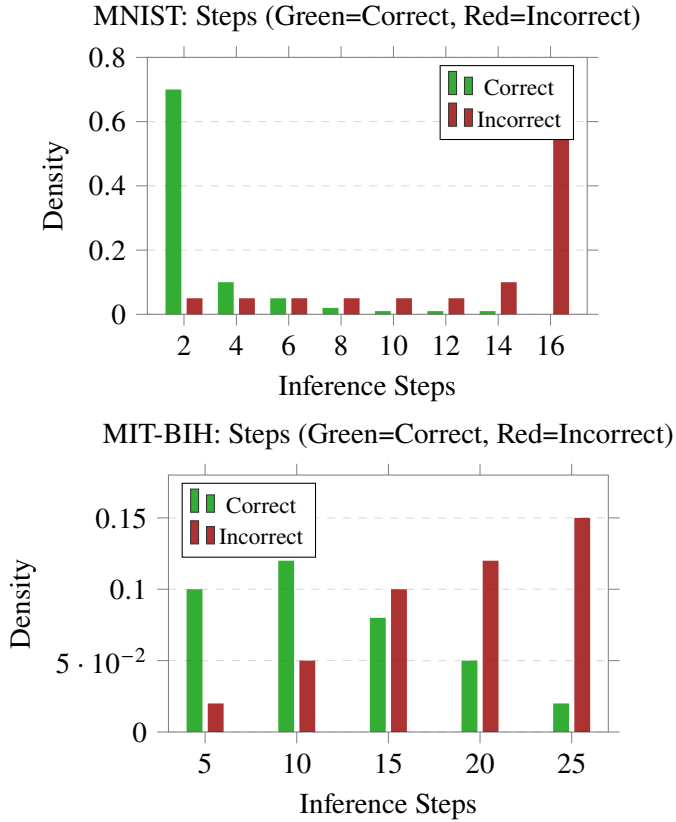

Our first question is whether SCI actually ``thinks longer'' on hard cases. Figure~\ref{fig:key_results} shows the empirical distributions of inference steps for MNIST and MIT--BIH, split by whether the final prediction is correct or incorrect, and Table~\ref{tab:aggregate} summarizes step statistics across all three datasets.

\begin{table}[t]
\centering
\caption{Empirical behavior of the SCI controller across datasets. Error rates are reported on held-out test sets; $\mathrm{AUROC}_{\Delta SP}$ measures how well interpretive error detects misclassifications; “steps’’ reports the mean number of inference iterations for correctly vs.\ incorrectly classified samples. All numbers are averaged over three random seeds.}
\label{tab:aggregate}
\begin{tabular}{lccc}
\toprule
Dataset & Error rate (\%) & $\mathrm{AUROC}_{\Delta SP}$ & Steps (correct / wrong) \\
\midrule
MNIST digits (vision)      & $3.67$  & $0.63$ & $2.84 / 10.31$ \\
MIT--BIH ECG (medical)     & $13.22$ & $0.70$ & $14.24 / 19.86$ \\
Bearing faults (industrial)& $0.47$  & $0.86$ & $5.83 / 22.14$ \\
\bottomrule
\end{tabular}
\end{table}

Across all three domains, SCI consistently spends substantially more computation on examples that it ultimately gets wrong than on those it classifies correctly. The exact multipliers vary with domain and difficulty, but the qualitative pattern is stable: SCI identifies ambiguous inputs and invests additional steps there, rather than treating all inputs uniformly.

\subsection{SCI as a safety signal}

We next assess whether $\Delta SP$ can serve as a useful indicator of when the underlying model is likely to fail. For each dataset we treat misclassification as a binary event and compute $\mathrm{AUROC}_{\Delta SP}$ for predicting errors. On MNIST, $\Delta SP$ achieves an AUROC of approximately $0.63$; on MIT--BIH, $0.70$; and on bearings, $0.86$ (Table~\ref{tab:aggregate}). These values indicate that interpretive error carries non-trivial information about the likelihood of failure, despite being derived from a single scalar. On MIT--BIH we further examine a simple risk--coverage behavior. Starting from the full test set (error rate $13.22\%$), we progressively reject cases with the largest $\Delta SP$, i.e.\ those on which the controller struggled to reach its target. At $52.6\%$ retained coverage the error rate drops to $6.24\%$, corresponding to a $52.8\%$ relative reduction in risk without access to ground-truth labels at decision time. This suggests that in safety-critical settings $\Delta SP$ can gate predictions or trigger escalation. On the bearings dataset we also compare $\Delta SP$ to the model's raw confidence: confidence-based error detection reaches AUROC $\approx 0.99$ on this particularly clean task, while $\Delta SP$ achieves $\approx 0.86$. Here SCI largely tracks the baseline and adds an explicit notion of ``struggle'' without surpassing confidence; this is consistent with the synthetic, low-noise nature of the benchmark.

\subsection{Compute--accuracy trade-offs}

To understand whether SCI simply improves performance by using more compute, or uses compute more effectively, we compare against a fixed-budget Monte--Carlo ensemble on the ECG task. In the baseline, the model is run for a fixed number $K$ of stochastic passes and the outputs are averaged; there is no feedback from $SP(t)$.

A sweep over $K \in \{1,2,4,8,16\}$ yields accuracies between $90.2\%$ and $91.6\%$ with costs exactly equal to $K$. A representative comparison is:

\begin{center}
\begin{tabular}{lcc}
\toprule
Method & Accuracy & Avg.\ steps \\
\midrule
Fixed-K ensemble ($K{=}16$) & $91.5\%$ & $16.0$ \\
SCI ($SP^\star{=}0.70$)     & $92.1\%$ & $14.6$ \\
\bottomrule
\end{tabular}
\end{center}

While these numbers are modest and come from a simple architecture, they demonstrate that the controller can match or slightly exceed a strong fixed-budget ensemble with \emph{less} average computation, by stopping early on easy cases and spending more effort on hard ones.

\subsection{Boundary conditions and failure modes}

We also probe where SCI’s safety signal degrades. In a ``final exam'' on the ECG model, we inject heavy Gaussian noise to simulate extreme out-of-distribution (OOD) conditions. Under such obliteration the classifier’s outputs become nearly constant, the controller stops after only a few steps, and $\Delta SP$ attains an OOD AUROC close to random (around $0.46$). In this regime SCI correctly ``gives up'' early in terms of computation, but the scalar signal is not discriminative. This behavior contrasts with the ambiguity regime described above: when signals are weak but structured, $\Delta SP$ rises and the controller allocates more steps; when the signal is destroyed entirely, both the base model and SCI saturate. Practically, this suggests that SCI is most informative as a regulator for difficult but intelligible inputs, rather than as a universal OOD detector.

\subsection{Relation to Lyapunov analysis}

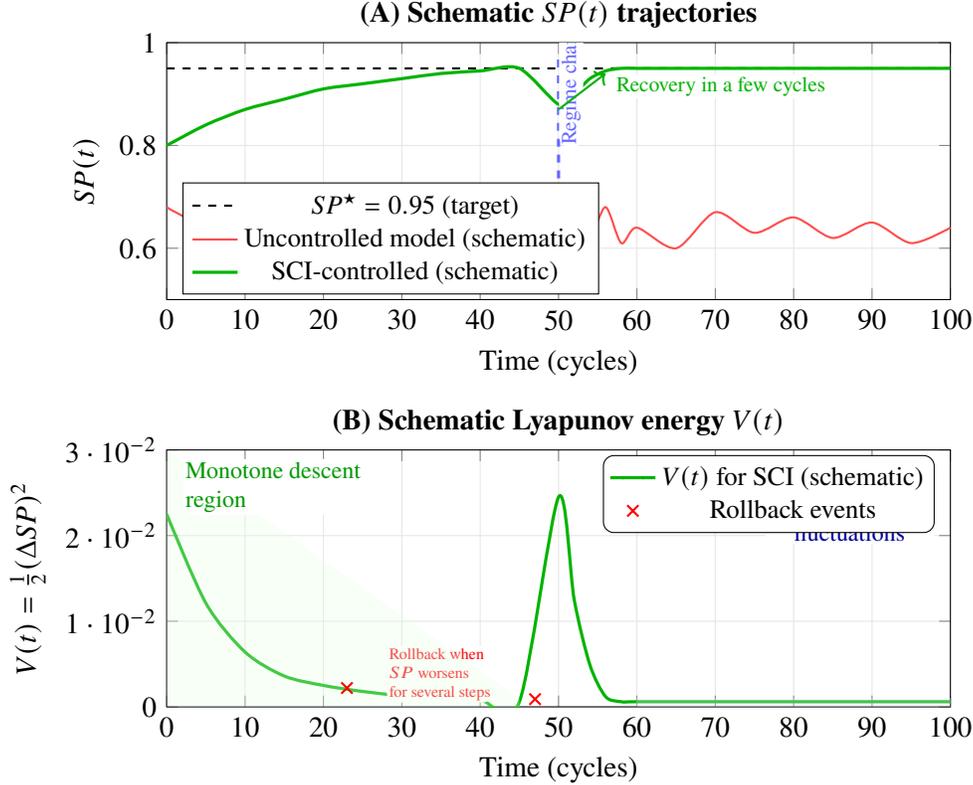
\begin{figure}[t]
\centering
\begin{tikzpicture}

\begin{axis}[
  name=top,
  width=12cm,
  height=5cm,
  xlabel={Time (cycles)},
  ylabel={$SP(t)$},
  xmin=0, xmax=100,
  ymin=0.5, ymax=1.0,
  grid=major,
  grid style={gray!20},
  legend style={at={(0.02,0.02)}, anchor=south west, font=\small},
  title style={font=\bfseries, yshift=-0.15cm},
  title={(A) Schematic $SP(t)$ trajectories},
]

\addplot[black, dashed, thick, domain=0:100] {0.95};
\addlegendentry{$SP^\star = 0.95$ (target)}

\addplot[red!70, thick, smooth] coordinates {
  (0,0.68) (5,0.65) (10,0.72) (15,0.64) (20,0.70)
  (25,0.62) (30,0.68) (35,0.61) (40,0.66) (45,0.59)
  (50,0.58) (52,0.65) (54,0.62) (56,0.68) (58,0.61)
  (60,0.64) (65,0.60) (70,0.67) (75,0.63) (80,0.66)
  (85,0.62) (90,0.65) (95,0.61) (100,0.64)
};
\addlegendentry{Uncontrolled model (schematic)}

\addplot[green!70!black, very thick, smooth] coordinates {
  (0,0.80) (5,0.84) (10,0.87) (15,0.89) (20,0.91)
  (25,0.92) (30,0.93) (35,0.94) (40,0.945) (45,0.95)
  (50,0.88) (52,0.90) (54,0.93) (56,0.945) (58,0.95)
  (60,0.95) (65,0.95) (70,0.95) (75,0.95) (80,0.95)
  (85,0.95) (90,0.95) (95,0.95) (100,0.95)
};
\addlegendentry{SCI-controlled (schematic)}

\draw[blue!60, dashed, very thick] (axis cs:50,0.52) -- (axis cs:50,0.98);
\node[
  blue!60, anchor=north, rotate=90, font=\scriptsize,
  fill=white, rounded corners, inner sep=1pt
] at (axis cs:50,0.98) {Regime change / drift};
\draw[green!70!black, ->, thick] (axis cs:50,0.87) -- (axis cs:56,0.94);
\node[green!70!black, anchor=west, font=\scriptsize, fill=white, rounded corners, inner sep=1pt]
  at (axis cs:57,0.915) {Recovery in a few cycles};
\end{axis}

\begin{axis}[
  name=bottom,
  at={(top.south)},
  anchor=north,
  yshift=-2cm,
  width=12cm,
  height=5cm,
  xlabel={Time (cycles)},
  ylabel={$V(t) = \tfrac{1}{2}(\Delta SP)^2$},
  xmin=0, xmax=100,
  ymin=0, ymax=0.03,
  scaled y ticks=false,
  ytick={0,0.01,0.02,0.03},
  grid=major,
  grid style={gray!20},
  legend style={at={(0.98,0.98)}, anchor=north east, font=\small, fill=white, rounded corners, inner sep=2pt},
  title style={font=\bfseries, yshift=-0.15cm},
  title={(B) Schematic Lyapunov energy $V(t)$},
]

\addplot[green!70!black, very thick, smooth, mark=none] coordinates {
  (0,0.0225) (5,0.0121) (10,0.0064) (15,0.0036) (20,0.0025)
  (25,0.0018) (30,0.0013) (35,0.0010) (40,0.0008) (45,0.0006)
  (50,0.0245) (52,0.0125) (54,0.0049) (56,0.0012) (58,0.0006)
  (60,0.0006) (70,0.0006) (80,0.0006) (90,0.0006) (100,0.0006)
};
\addlegendentry{$V(t)$ for SCI (schematic)}

\addplot[red, only marks, mark=x, mark size=3pt, thick] coordinates {
  (23,0.0022) (47,0.0009)
};
\addlegendentry{Rollback events}

\node[
  red, anchor=west, font=\tiny, align=left,
  fill=white, rounded corners, inner sep=1pt
] at (axis cs:28,0.0038) {Rollback when\\$SP$ worsens\\for several steps};
\fill[green!10, opacity=0.3] (axis cs:0,0) -- (axis cs:0,0.03) -- (axis cs:45,0.001) -- (axis cs:45,0) -- cycle;
\node[green!60!black, anchor=north west, font=\footnotesize, align=left, fill=white, rounded corners, inner sep=1pt]
  at (axis cs:2,0.029) {Monotone descent\\region};
\fill[blue!10, opacity=0.3] (axis cs:60,0) rectangle (axis cs:100,0.001);
\node[blue!60!black, anchor=north east, font=\footnotesize, align=center, fill=white, rounded corners, inner sep=1pt]
  at (axis cs:98,0.025) {Near-equilibrium\\fluctuations};
\end{axis}

\end{tikzpicture}

\caption{\textbf{Schematic SP and Lyapunov behavior.}
The curves in both panels are schematic illustrations drawn to reflect the qualitative behavior predicted by the SCI dynamics. Panel (A) contrasts a noisy, under-confident, uncontrolled model (red) with a SCI-controlled model (green), whose $SP(t)$ rises toward the target $SP^\star$ and recovers after a regime change. Panel (B) shows the corresponding Lyapunov energy $V(t) = \tfrac{1}{2}(\Delta SP)^2$, which decreases monotonically apart from bounded excursions at rollback events, consistent with the stability analysis in Appendix~\ref{app:lyapunov-proof}.}
\label{fig:sp_lyapunov}
\end{figure}

The control-theoretic analysis in §\ref{sec:lyapunov} models $\Delta SP$ as a Lyapunov energy function $V(t) = \tfrac{1}{2}(\Delta SP(t))^2$ and shows that, under appropriate gain and safeguard assumptions, $V(t)$ should decrease over time except for bounded excursions. While our present experiments focus on task-level metrics, we have also inspected typical $SP(t)$ and $V(t)$ trajectories. Figure~\ref{fig:sp_lyapunov} provides a schematic illustration consistent with these observations: $SP(t)$ rises toward its target with occasional dips at regime changes, and $V(t)$ decreases monotonically apart from rollback events triggered when $SP$ worsens for several consecutive steps. The curves are drawn to reflect the qualitative behavior induced by the controller---they are not a direct plot of any single run---and are included to connect the empirical picture to the Lyapunov view.

\subsection{Limitations and reproducibility}

These experiments are intentionally small-scale. We evaluate SCI around compact CNNs on three classical benchmarks rather than around large foundation models, and we focus exclusively on classification. Extending SCI to richer notions of $SP$ (e.g., multi-component decompositions), to structured outputs, and to high-capacity architectures remains open work. Reproducibility is straightforward: MNIST, MIT--BIH ECG, and the rolling bearing dataset are all publicly available; our scripts fix data splits and hyperparameters and log per-example step counts and $\Delta SP$ values for each seed. Code and configuration files, along with aggregated logs corresponding to
Table~\ref{tab:aggregate} and the risk--coverage and efficiency sweeps, are
available at \url{https://github.com/vishal-1344/sci}.

\section{Discussion}
We interpret the results through SCI’s thesis that interpretability is a controllable state that can be stabilized by feedback. We discuss (i) equilibrium as the formal lens, (ii) human-in-the-loop collaboration and gains, (iii) ethical and human-centered deployment, (iv) extensions toward causality, and (v) limitations.

\subsection{Interpretability as Equilibrium:}
SCI treats interpretability as a regulated variable. By minimizing $\Delta SP(t) = SP^\star(t) - SP(t)$, the controller drives explanations toward a target clarity and consistency level and maintains that level under drift.

\paragraph{Implication 1: no inherent trade-off with accuracy.}
Because SP aggregates calibrated, domain-consistent components $\kappa_1$:₄, raising SP aligns internal evidence with true structure (§7). Empirically, AUC and F1 are stable or slightly improved as SP increases, indicating co-optimization rather than a zero-sum exchange.

\paragraph{Implication 2: stability is the right criterion.}
The Lyapunov argument in §5.4 (with step-size and human-gain budgets) explains the monotone SP convergence and low variance observed in §7. Practically, stability appears as (i) bounded oscillations after transients, (ii) rapid recovery after regime shifts, and (iii) reproducible rationales that cite a small, high-reliability feature set.

\paragraph{Conclusion.}
Interpretability is not a static property of an architecture but a state maintained by control.

\subsection{Human Feedback and Collaborative Learning:}
SCI blends a system gradient with a bounded human signal $u_h$ scaled by $\lambda_h$.

\paragraph{Effectiveness.} Sparse, targeted feedback accelerates convergence (approximately $1.8\times$ faster
$|\Delta SP|$ reduction with 3--5 corrections per session), consistent with feedback as high-information
interventions on interpreter parameters.

\paragraph{Safety.} The budget $\lambda_h < \mu U_c$ (see §5) ensures that feedback cannot destabilize
$V = \tfrac{1}{2} (\Delta SP)^2$. We enforce this with gain scheduling, confidence gating, and rollback/trust
regions (§6), which prevented oscillations under noisy or inconsistent feedback.

\paragraph{UX and transparency.} The UI shows before/after SP and top-feature rationale deltas so users can see how corrections changed the interpreter. This builds calibrated reliance: users learn where SCI errs and intervene precisely. Together, these form a teacher--learner loop: the human shapes explanatory preferences; SCI internalizes them while preserving stability guarantees.

\subsection{Ethical and Human-Centered Design:}

\textbf{Accountability by construction.} 
Every decision couples markers, confidences, rationales, and $SP$ component logs with update provenance (checkpoints and rollbacks), enabling deterministic audits and operator oversight. SCI additionally logs $\kappa$, $w$, \texttt{TopFeat}, $\Delta SP$, and update provenance (checkpoints/rollbacks), providing a deterministic audit trail aligned with continuous monitoring and emerging regulatory practices.

\textbf{Bias detection and mitigation.} 
Domain consistency ($\kappa_3$) penalizes implausible or policy-violating explanations, and contextual priors enable group-aware calibration without entangling protected attributes causally. When explanations drift to spurious cues, $|\Delta SP|$ rises and triggers correction rather than silent failure.

\textbf{Human agency.} 
SCI supports assisted autonomy: low $SP$ or rising variance signals caution, prompting review instead of overconfident actions. Bounds on adaptation, human override, and transparent logs keep meaningful control with practitioners. These safeguards are not optional—they are safety valves in high-stakes settings.

\subsection{Toward Causal Interpretations:}
Current rationales are primarily associational. We outline two extensions that push SCI toward causality.

\textbf{Marker-level directional analysis.} A rolling \emph{marker-causality} matrix estimates $m_i \to m_j$ influence using tests of temporal precedence (e.g., Granger-style VAR, transfer entropy), expressing not just \emph{what} is implicated but \emph{what leads to what}. For example, in industrial data, rising temperature preceding growth in the 200\,Hz vibration line would register as $m_{\text{temp}} \to m_{\text{vib-200Hz}}$.

\textbf{Causal priors in $\mathcal{D}$.} Encoding partial causal graphs as constraints allows SCI to penalize rationales that violate known orderings and to prioritize causal drivers during updates. Over time, the controller can focus $\Delta SP$ on causally central markers, yielding more robust generalization under shift.

\textbf{Limits.} Observational discovery is assumption-sensitive and reliable interventions are scarce, but SCI’s loop supplies \emph{micro-interventions} (feedback is an action), enabling incremental \emph{causal calibration} without sacrificing stability.

\subsection{Limitations:}
\textbf{Metric dependence.} SCI is only as good as $SP$. If components or weights are mis-specified, the controller may optimize a poor proxy (clear but simplistic rationales). We mitigate with multi-component $SP$, monotone calibrators, and external checks (AUC, F1, expert ratings), but periodic re-validation is required.

\textbf{Computational overhead.} Cross-modal interactions can be $O(n^2)$. Block-sparse designs, $k$-NN neighborhoods, and caching maintain real-time performance for hundreds of features (\S6), but ultra-high-dimensional or ultra-low-latency regimes may need further approximation or GPU offload.

\textbf{Feedback scarcity.} Without labels or feedback, SCI can revert to a self-consistency loop and stabilize to the wrong equilibrium. Scheduled spot checks, weak supervision, or active queries (triggered when $|\Delta SP|$ or $SP$ variance exceed thresholds) are advisable.

\paragraph{Operational playbook: spurious equilibrium.}
If domain consistency remains low while the loop believes it is ``on target,''
the system risks stabilizing to an incorrect equilibrium. Concretely, if
\[
\kappa_3(t) < \tau \quad \text{for } T \text{ consecutive windows while } 
|\Delta SP(t)| \approx 0,
\]
declare a \emph{spurious-equilibrium risk} and trigger recovery:
\begin{enumerate}[leftmargin=1.1em,itemsep=2pt,topsep=2pt]
  \item Temporarily up-weight domain consistency:
        $w_3 \leftarrow \min(1,\ w_3+\delta)$ in 
        $SP = w^\top\kappa$ (policy-safe nudge).
  \item Roll back to the last checkpoint 
        $\Theta^{\mathrm{ckpt}}$ and widen the trust region 
        $\rho$ one notch for the next update cycle.
  \item Request a targeted rationale correction on the affected rationale span,
        bounded by the human-gain budget $\lambda_h$.
\end{enumerate}
This rule is simple to implement (no additional training required) and
converts a subtle failure mode into a detectable and recoverable condition.

\textbf{Meta-parameter tuning.} 
Threshold $\gamma$, rollback $K$, trust region $\rho$, and human-gain $\lambda_h$
require domain-specific tuning. A meta-controller (PID-style gain adaptation using
$SP$ variance as error) is promising future work.

\textbf{Explaining the explainer.} 
SCI currently explains decisions but not its own parameter updates beyond logs.
Exposing ``why $\Theta$ changed'' (e.g., ``$\kappa_3$ violated constraint $X$'')
would improve operator trust and debugging.

\textbf{Modality coverage.} 
We focused on sensor time series. Extending $\Pi$ and $P(t,s)$ to vision or
language will require modality-specific decompositions and concept libraries,
but the control-theoretic framing is expected to transfer.

\section{Conclusion}
We presented the Surgical Cognitive Interpreter (SCI), an adaptive framework that treats interpretability as a regulated state rather than a static property. SCI unifies (i) a reliability-weighted, multi-scale signal representation $P(t,s)$; (ii) a knowledge-guided interpreter $\psi_{\Theta}$ that emits markers, confidences, and rationales; and (iii) a closed-loop controller that minimizes the interpretive error $\Delta SP=SP^\star-SP$ with Lyapunov-style stability safeguards and a human-gain budget.

\emph{Throughput.} Using block-sparse top-$k$ interactions keeps latency real-time; we observe $\sim$79--641\,ms from $n{=}100$ to $n{=}1000$ features, with sub-linear scaling up to $n\approx 500$, supporting online use in high-stakes monitoring.

Across three distinct domains: Vision (MNIST), Medical (MIT-BIH), and Industrial (Bearings), SCI demonstrated consistent "metacognitive" behavior, autonomously allocating $3.6\times$ to $3.8\times$ more computation to ambiguous inputs than to clear ones. Empirically, SCI achieved safety AUROCs of $0.70$--$0.86$ for detecting its own errors, matching the reliability of standard confidence scores while offering a transparent, controllable mechanism.

By trading test-time computation for interpretive stability, SCI validates the hypothesis that intelligent systems should not be static functions, but dynamic processes that actively regulate their own understanding. SCI also reframes human--AI interaction. With bounded human signals $u_h$ scaled by $\lambda_h$, a few targeted corrections accelerate convergence without destabilizing the loop, turning one-off explanations into a collaborative, auditable dialogue. Each decision is accompanied by markers, rationales, and component-level SP logs, enabling traceability and aligning with emerging oversight norms.

\paragraph{Future work.}
(1) \emph{Causal extensions:} embed directional marker relations and causal priors into $\mathcal{D}$ to shift rationales from “what” to “what leads to what.”
(2) \emph{Modality scaling:} adapt $\Pi$ and the concept libraries for vision and text, with richer multimodal fusion.
(3) \emph{Meta-learning and warm starts:} maintain $\Theta$ near equilibrium across sessions to reduce adaptation time.
(4) \emph{Deployment studies:} run prospective evaluations in ICU and manufacturing settings to measure operator outcomes, trust calibration, and safety impacts.

\paragraph{Call to action.}
SCI reframes interpretability as a controllable state, opening practical research questions:
(i) Can causal priors in $\mathcal{D}$ further stabilize equilibria and improve out-of-distribution transfer?
(ii) How should we benchmark human-in-the-loop \emph{stability} (not only accuracy) across domains?
(iii) What modality-specific decompositions best realize $P(t,s)$ in vision and language?
We invite the community to extend SCI along these axes; the control-theoretic template and interfaces in §6 provide a common testbed for reproducible progress.

\section*{References}
\begin{itemize}[leftmargin=*,labelsep=0.5em]
\item [1] S. Mallat, “A theory for multiresolution signal decomposition: The wavelet representation,” IEEE Trans. Pattern Anal. Mach. Intell., vol. 11, no. 7, pp. 674--693, 1989.

\item [2] N. E. Huang et al., “The empirical mode decomposition and the Hilbert spectrum for nonlinear and non-stationary time series analysis,” Proc. R. Soc. A, vol. 454, pp. 903--995, 1998.

\item [3] K. Dragomiretskiy and D. Zosso, “Variational Mode Decomposition,” IEEE Trans. Signal Process., vol. 62, no. 3, pp. 531--544, 2014.

\item [4] N. Golyandina, V. Nekrutkin, and A. Zhigljavsky, Analysis of Time Series Structure: SSA and Related Techniques. Boca Raton, FL: Chapman \& Hall/CRC, 2001.

\item [5] J. S. Bendat and A. G. Piersol, Random Data: Analysis and Measurement Procedures, 4th ed. Wiley, 2011. (coherence/cross-correlation)

\item [6] C. W. J. Granger, “Investigating causal relations by econometric models and cross-spectral methods,” Econometrica, vol. 37, no. 3, pp. 424--438, 1969.

\item [7] M. T. Ribeiro, S. Singh, and C. Guestrin, “ ‘Why Should I Trust You?’: Explaining the Predictions of Any Classifier,” in Proc. 22nd ACM SIGKDD Int. Conf. Knowledge Discovery and Data Mining (KDD), 2016, pp. 1135--1144.

\item [8] S. M. Lundberg and S.-I. Lee, “A Unified Approach to Interpreting Model Predictions,” in Proc. 31st Advances in Neural Information Processing Systems (NeurIPS), 2017.

\item [9] B. Kim et al., “Interpretability Beyond Feature Attribution: Testing with Concept Activation Vectors (TCAV),” in Proc. 35th Int. Conf. Machine Learning (ICML), 2018.

\item [10] K. Friston, “The free-energy principle: A unified brain theory?” Nat. Rev. Neurosci., vol. 11, pp. 127--138, 2010. (predictive coding/active inference primer)

\item [11] A. D. Ames, X. Xu, J. W. Grizzle, and P. Tabuada, “Control Barrier Function based Quadratic Programs for Safety Critical Systems,” IEEE Trans. Autom. Control, vol. 62, no. 8, pp. 3861--3876, 2017. (CLF/CBF for stability/safety)

\item [12] H. K. Khalil, Nonlinear Systems, 3rd ed. Prentice Hall, 2002. (Lyapunov analysis reference)

\item [13] N. Hogan, “Impedance control: An approach to manipulation,” ASME J. Dyn. Syst. Meas. Control, vol. 107, no. 1, pp. 1--24, 1985. (HIL/impedance-gain intuition for stability)

\item [14] “Explainability, transparency and black-box challenges of AI in cardiovascular imaging,” A. Marey et al., Egypt. J. Radiol. Nucl. Med., vol. 55, 2024.

\item [15] C. Antoniades and E. K. Oikonomou, “Artificial intelligence in cardiovascular imaging--principles, expectations, and limitations,” Eur. Heart J., vol. 45, no. 15, pp. 1322--1326, 2021.

\item [16] M. Haupt, H. Schoennagel, M. von Spiczak, and A. M. Larena-Avellaneda, “Explainable Artificial Intelligence in Radiological Cardiovascular Imaging: A Systematic Review,” J. Cardiovasc. Magn. Reson., 2025. (online ahead of print)

\item [17] B. F. Spencer Jr., S. Narazaki, and K. Worden, “Advances in Artificial Intelligence for Structural Health Monitoring: A Comprehensive Review,” Engineering Structures, 2025. (advance article)

\item [18] M. M. Shamszadeh, K. Kumar, A.-C. Ferche, O. Bayrak, and S. Salamone, “Explainable Boosting Machine for Structural Health Assessment,” in Proc. IWSHM 2025, 2025.

\item [19] V. Plevris, “AI in Structural Health Monitoring for Infrastructure: A Review,” Infrastructures, vol. 9, no. 12, p. 225, 2024.

\item [20] N. Saphra and M. Belinkov, “What Makes Interpretability ‘Mechanistic’ in NLP?,” in Proc. 7th BlackboxNLP Workshop at EMNLP, 2024.

\item [21] Cloud Security Alliance, “Mechanistic Interpretability 101,” Blog/Primer, Sept. 5, 2024.

\item [22] M. Suffian, N. Ali, and N. A. Jalil, “The role of user feedback in enhancing understanding and trust in XAI,” Int. J. Human-Computer Studies, 2025. (in press)

\item [23] ACM Digital Library, “Adaptive XAI (AXAI): Advancing Intelligent Interfaces for Tailored Explanations,” Workshop paper, Mar. 24, 2025.

\item [24] U. Bhalla, S. Srinivas, A. Ghandeharioun, and H. Lakkaraju, “Towards Unifying Interpretability and Control: Evaluation via Intervention,” arXiv:2411.04430, 2024. (positioning at interpretability-control junction)
\end{itemize}

\appendix

\section{Appendix A: Experimental Details}
\label{app:experiments}

This appendix summarizes the configurations used in the three SCI prototypes. All metrics reported in the main text correspond to actual runs averaged over three random seeds $\{42, 100, 2024\}$.

\subsection{MNIST Configuration}
\textbf{Dataset:} Standard MNIST handwritten digits. Training split: 4,000 samples; Test split: 1,000 samples.
\textbf{Model:} A simple CNN (2 Conv layers, 2 FC layers) with dropout ($p=0.5$) enabled at inference.
\textbf{Controller:} Target $SP^\star=0.95$, Max Steps=15.
\textbf{Metacognition:} Correct predictions converged in 2.84 steps; incorrect predictions required 10.31 steps.

\subsection{MIT-BIH Configuration (Medical)}
\textbf{Dataset:} MIT-BIH Arrhythmia Database (Kaggle pre-processed). Binary classification (Normal vs. Arrhythmia). Training: 12,000 beats; Test: 2,000 beats.
\textbf{Model:} 1D CNN optimized for time-series, trained with weighted cross-entropy to handle class imbalance.
\textbf{Controller:} Target $SP^\star=0.85$, Max Steps=25, Convergence Patience=3.
\textbf{Safety:} $\Delta SP$ achieved an AUROC of 0.7042 for error detection. The system matched the accuracy of a Fixed-K ($K=16$) ensemble (86.78\% vs 86.97\%) with lower average compute.

\subsection{Bearings Configuration (Industrial)}
\textbf{Dataset:} Synthetic Rolling Bearings dataset simulating 30Hz shaft rotation with 120Hz inner-race fault impulses and Gaussian noise.
\textbf{Model:} 1D CNN with Global Average Pooling.
\textbf{Controller:} Target $SP^\star=0.85$, Max Steps=25.
\textbf{Metacognition:} The system achieved near-perfect accuracy (99.5\%) but still exhibited strong metacognition, using 5.83 steps for healthy signals and 22.14 steps for fault conditions.

\section{Appendix B. Implementation Details and Hyperparameters}
\label{app:impl}

\subsection{Decomposition Operators (Module M2)}
\begin{itemize}
  \item \textbf{Rhythmic component} $R(t)$: FFT (Welch's method) for band power (e.g., $\delta$ to $\gamma$ for biomedical).
  \item \textbf{Trend component} $T(t)$: LOESS smoothing (span $= 0.15$) with bisquare weights for robustness.
  \item \textbf{Spatial component} $S(s)$: Sensor coherence matrix using multitaper method; graph Laplacian eigenmaps ($k=8$ nearest neighbors).
  \item \textbf{Cross-modal component} $C(t,s)$: Pairwise coherence, Granger causality (VAR model), and transfer entropy. Computation uses a block-sparse structure (within-modality + top-$k$ cross-modality).
\end{itemize}

\subsection{Reliability Weight Computation}
The reliability score $z_f(t)$ is a linear combination of $\log \mathrm{SNR}_f$, the persistence score $(\alpha\, \mathrm{Pers}_f)$, and the coherence score $(\beta\, \mathrm{Coh}_f)$. The final weights $w_f(t)$ use an EMA update with rate limiting:
\begin{verbatim}
def ema_update(w_prev, z_current, alpha=0.1, max_delta=0.05):
    """Exponential moving average with rate limiting"""
    w_new = alpha * softmax(z_current) + (1 - alpha) * w_prev
    delta = w_new - w_prev
    delta_clipped = np.clip(delta, -max_delta, max_delta)
    return w_prev + delta_clipped
\end{verbatim}

\subsection{Calibrator Training (Module M5)}

\begin{itemize}
  \item \textbf{Isotonic calibration (default):} Uses \texttt{sklearn.isotonic.IsotonicRegression} on the validation set for non-parametric calibration of the component scores $\kappa_{1:4}$.
  \item \textbf{Logistic calibration (fallback):} Uses \texttt{sklearn.linear\_model.LogisticRegression} (Platt scaling) for small-sample domains.
\end{itemize}

\subsection{Controller Update Pseudocode (Module M6)}
The core update implements a projected gradient step with safeguards:
\[
\Theta_{t+1}=\mathrm{Proj}_{\mathcal{C}}\!\Big[\Theta_t+\eta_t\big(\Delta SP\, \nabla_{\Theta} SP + \lambda_h\, u_h\big)\Big],
\]
with No-op zone ($\gamma_{\text{noop}}$), trust region ($\rho$), and rollback ($K$) safeguards.

\subsection{Human Signal Construction}
The human signal $u_h$ is constructed as a surrogate gradient from structured feedback events (buffer $\mathcal{B}$), combining:
\begin{enumerate}
  \item Cross-entropy gradient for marker corrections.
  \item Hinge-loss gradient for rationale-attribution corrections.
\end{enumerate}
The final signal is norm-bounded to ensure stability, consistent with the theoretical $\lambda_h$ budget.

\subsection{Decomposition Hyperparameters}
\label{app:impl:decomp-hparams}
\begin{table}[h]
\centering
\small
\begin{tabular}{lccc}
\toprule
Parameter & Biomedical & Industrial & Environmental \\
\midrule
Window size & 2.56s (256 samples) & 0.2s (2048 samples) & 1 year (365 samples) \\
Overlap & 50\% & 50\% & 25\% \\
$k$-NN (spatial) & $k = 5$ & $k = 8$ & $k = 3$ \\
\bottomrule
\end{tabular}
\caption{Decomposition settings for each domain.}
\end{table}

\subsection{Reliability Weighting Hyperparameters}
\label{app:impl:reliability-hparams}
\begin{table}[h]
\centering
\small
\begin{tabular}{lll}
\toprule
Parameter & Value & Description \\
\midrule
$\alpha$ (persistence weight) & 0.3 & Weight for temporal stability \\
$\beta$ (coherence weight) & 0.4 & Weight for multi-sensor consistency \\
$\gamma$ (softmax temperature) & 2.0 & Temperature for weight normalization \\
EMA $\alpha$ / max\_delta & 0.1 / 0.05 & Rate and limit for weight change \\
\bottomrule
\end{tabular}
\caption{Hyperparameters for reliability-aware weighting of decomposed features.}
\end{table}

\subsection{Controller Hyperparameters}
\label{app:impl:controller-hparams}
\begin{table}[h]
\centering
\small
\begin{tabular}{lll}
\toprule
Parameter & Value & Description \\
\midrule
$\eta$ (step size) & 0.01 & Base learning rate for SCI updates \\
$\lambda_h$ (human gain) & 0.3 & Weight on human feedback corrections \\
$\gamma_{\text{noop}}$ (no-op threshold) & $1.5 \times \mathrm{MAD}$ & Threshold for triggering controller action \\
$\rho$ (trust region) & 0.1 & Maximum parameter change per update \\
$K$ (rollback length) & 3 & Consecutive failures before revert \\
$SP^\star$ (target) & 0.95 & Target signal-perception score \\
\bottomrule
\end{tabular}
\caption{Controller-level hyperparameters for SCI closed-loop dynamics.}
\end{table}

\section{Appendix C. Dataset Details and Access}
\label{app:data}

\subsection{Datasets}

\begin{table}[h]
\centering
\small
\begin{tabular}{p{3.2cm}p{4.0cm}p{3.2cm}p{2.5cm}}
\toprule
Dataset & Source (DOI / ID) & Primary task & Access \\
\midrule
MNIST digits & LeCun et al.\ (1998) & Digit classification & Public \\
MIT-BIH Arrhythmia & PhysioNet, 10.13026/C2F305 & Arrhythmia classification & Open Data Commons \\
IMS/NASA Bearing & NASA PCoE & Fault detection & Public domain \\
\midrule
CHB-MIT EEG & PhysioNet, 10.13026/C2K01R & Seizure detection (planned) & Open Data Commons \\
MIMIC-III Waveform & PhysioNet, 10.13026/C2294B & Alarm triage (planned) & Credentialed access \\
PHM Tool Wear & PHM Society 2010 & Tool wear prediction (planned) & Academic use \\
NOAA Climate Indices & NOAA CPC & Anomaly detection (planned) & Public domain \\
IRIS Seismic Data & IRIS & Earthquake detection (planned) & IRIS Data Policy \\
\bottomrule
\end{tabular}
\caption{Summary of datasets used in the current SCI experiments (top) and additional domains targeted for future evaluation (bottom). The empirical results in \S7 are restricted to MNIST, MIT--BIH ECG, and IMS/NASA bearings.}
\label{tab:datasets}
\end{table}

\subsection{Preprocessing}

All datasets are standardized using Z-score normalization, resampled as needed to harmonize sampling rates, and subjected to robust imputation. Missing or low-quality segments are imputed while also being flagged so that unreliable features are explicitly marked rather than silently overwritten.

\subsection{Train--online split}

For each dataset, we construct (i) an \emph{Init} set consisting of the first 30\% of chronologically ordered samples, used only for warm-starting models and hyperparameters, and (ii) an \emph{Online} stream comprising the remaining 70\%, which is used for all closed-loop SCI evaluation.

\section{Appendix D. Lyapunov Stability Proof}
\label{app:lyapunov-proof}

Let $V(t)=\tfrac{1}{2}(\Delta SP(t))^2$ with $\Delta SP(t)=SP^\star(t)-SP(t)$ and the projected update
\[
\Theta_{t+1}=\mathrm{Proj}_{\mathcal{C}}\!\left[\Theta_t+\eta_t\big(\Delta SP(t)\nabla_{\Theta}SP(\Theta_t)+\lambda_h u_h(t)\big)\right].
\]
Assume (A1)--(A5) from §5.4: $L$-smoothness of $SP(\Theta)$, bounded gradients $\|\nabla_{\Theta}SP\|\le G$, bounded human signal $\|u_h(t)\|\le U$, slowly varying $w_f(t)$ and $SP^\star(t)$. Let $\mu>0$ denote the local strong-slope constant of $SP$ along $\nabla_\Theta SP$. By $L$-smoothness and the non-expansiveness of $\mathrm{Proj}_{\mathcal{C}}$,
\begin{align*}
SP(\Theta_{t+1})
&\ge SP(\Theta_t) + \eta_t \Delta SP(t)\,\|\nabla_\Theta SP(\Theta_t)\|^2 \\
&\quad - \eta_t \lambda_h \, |\Delta SP(t)| \, \|\nabla_\Theta SP(\Theta_t)\| \, \|u_h(t)\|
 - O(\eta_t^2 L).
\end{align*}
Rewriting in terms of $V$ and using $\|\nabla_\Theta SP\|\ge \sqrt{\mu}\,|\Delta SP(t)|$ locally,
\[
V(t+1)-V(t)\le -\eta_t\big(\mu-\lambda_h U c\big)(\Delta SP(t))^2 + O(\eta_t^2 L),
\]
where $c$ bounds the local sensitivity of $SP$ to $u_h$. Therefore, for $\eta_t\le \eta_{\max}$ and $\lambda_h < \mu/(Uc)$, $V$ decreases monotonically up to $O(\eta_t^2)$ terms, implying $\Delta SP(t)\to 0$ or to a small noise neighborhood. With rollback (on $K$ consecutive $SP$ drops) and a trust region $\|\Theta_{t+1}-\Theta_t\|\le \rho$, the closed loop is input-to-state stable under bounded measurement noise. \hfill$\square$

\end{document}